\documentclass[nohyperref]{article}

\usepackage{microtype}
\usepackage{graphicx}
\usepackage{subfigure}
\usepackage{booktabs} %

\usepackage{hyperref}

\usepackage[accepted]{icml2022}

\usepackage{amsmath}
\usepackage{amssymb}
\usepackage{mathtools}
\usepackage{amsthm}

\usepackage[capitalize,noabbrev]{cleveref}

\theoremstyle{plain}
\newtheorem{theorem}{Theorem}[section]

\theoremstyle{definition}
\newtheorem{definition}[theorem]{Definition}

\theoremstyle{remark}

\usepackage[textsize=tiny]{todonotes}

\usepackage{array, arydshln}

\usepackage{algorithm}
\usepackage{algorithmic}
\usepackage{xcolor}

\definecolor{green}{RGB}{11,155,13}

\newcommand{\ind}{\perp\!\!\!\!\perp}
\newcommand{\notind}{\not\!\perp\!\!\!\perp}
\newcommand{\dS}{{d_\mathcal{S}}}

\usepackage{bbm}
\usepackage{diagbox}
\usepackage{dsfont}
\usepackage{makecell}
\usepackage[algo2e, ruled, linesnumbered]{algorithm2e}
\hypersetup{
    colorlinks=true,
    linkcolor=black,
    filecolor=magenta,      
    urlcolor=cyan
    }

\DeclareMathOperator*{\argmin}{arg\,min}
\DeclareMathOperator*{\argmax}{arg\,max}
\usepackage{multirow}
\usepackage{etoolbox}
\makeatletter
\patchcmd{\@makecaption}
  {\scshape}
  {}
  {}
  {}
\makeatother

\begin{document}

\twocolumn[
\icmltitle{Causal Dynamics Learning for Task-Independent State Abstraction}

\icmlsetsymbol{equal}{*}

\begin{icmlauthorlist}
\icmlauthor{Zizhao Wang}{utece}
\icmlauthor{Xuesu Xiao}{utcs}
\icmlauthor{Zifan Xu}{utcs}
\icmlauthor{Yuke Zhu}{utcs}
\icmlauthor{Peter Stone}{utcs,sony}
\end{icmlauthorlist}

\icmlaffiliation{utece}{Department of Electrical and Computer Engineering, }
\icmlaffiliation{utcs}{Department of Computer Science, The University of Texas at Austin, Austin, USA}
\icmlaffiliation{sony}{Sony AI}

\icmlcorrespondingauthor{Zizhao Wang}{zizhao.wang@utexas.edu}

\icmlkeywords{Machine Learning, ICML}

\vskip 0.3in
]

\printAffiliationsAndNotice{}  %

\begin{abstract}
Learning dynamics models accurately is an important goal for Model-Based Reinforcement Learning (\textsc{mbrl}), but most \textsc{mbrl} methods learn a \textit{dense} dynamics model which is vulnerable to spurious correlations and therefore generalizes poorly to unseen states.
In this paper, we introduce \textit{Causal Dynamics Learning for Task-Independent State Abstraction} (\textsc{cdl}), which first learns a theoretically proved \textit{causal} dynamics model that removes unnecessary dependencies between state variables and the action, thus generalizing well to unseen states.
A state abstraction can then be derived from the learned dynamics, which not only improves sample efficiency but also applies to a wider range of tasks than existing state abstraction methods.
Evaluated on two simulated environments and downstream tasks, both the dynamics model and policies learned by the proposed method generalize well to unseen states and the derived state abstraction improves sample efficiency compared to learning without it.
\end{abstract}

\section{Introduction}
\label{sec:intro}

\begin{figure}
  \centering
  \includegraphics[width=1.0\columnwidth]{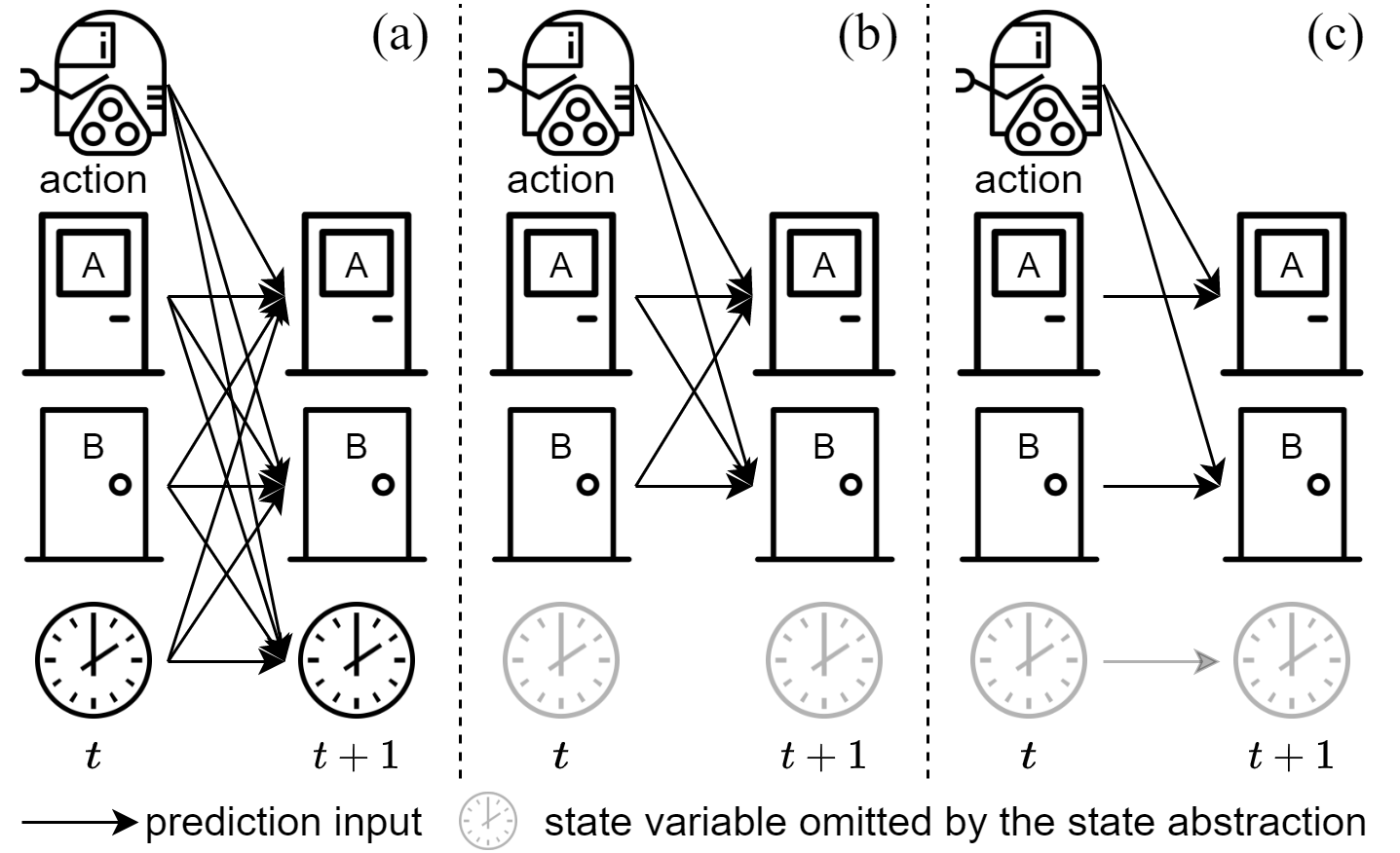}
  \vspace{-20pt}
  \caption{\small With two doors that the robot can open and go through and a clock on the wall, (a) \emph{dense} dynamics models predict dynamics of each state variable unnecessarily using all variables; (b) based on a pre-defined reward (e.g., for navigation), existing state abstractions learn to omit the clock but still use a dense model for the remaining variables; (c) our \emph{causal} models reason and only keep necessary dependencies (i.e., doors A and B depend on the action individually) and derives a state abstraction independent from any reward function.}
  \label{fig:method_comparison}
  \vspace{-20pt}
\end{figure}

\emph{Model-based Reinforcement Learning} (\textsc{mbrl}) enables an agent to predict what would happen if it executed various actions, thus allowing it to learn from \emph{imagined} experience \cite{hafner2019learning}. However, such an approach relies on the model being learned \emph{accurately}. Many model-based approaches rely on \emph{dense} models that predict the next step value of each variable based on the action and all variables in the current state, as shown in Fig. \ref{fig:method_comparison} (a). Such dense models are sensitive to spurious correlations which lead to poor generalization. For example, when door B is at angles unseen during training or the clock is at unseen times, the prediction of door A can be inaccurate due to unnecessary dependence on those variables in the model.

Observing that unnecessary dependencies are the source of poor generalization, existing state abstractions mitigate this problem by removing some of those dependencies. Specifically, state abstractions group many states into an abstract state by omitting some state variables \cite{chapman1991input, mccallum1996reinforcement, jong2005state}.  For example, bisimulation \cite{zhang2020learning}, which is particularly related to this paper, omits variables irrelevant to a pre-defined reward function. Though doing so removes unnecessary dependence on omitted variables, as shown in Fig. \ref{fig:method_comparison} (b), the same generalization issues persist as dense models are still used for the remaining variables, leaving unnecessary dependencies in the abstract state. Furthermore, despite bisimulation improving sample efficiency of task learning by reducing the learning space, such methods only find problem-dependent abstractions: they may omit variables that are irrelevant to the current task but that the agent can control and utilize for future tasks.

Given that good generalization requires only keeping necessary dependencies, this paper introduces Causal Dynamics Learning  (\textsc{cdl}) for Task-Independent State Abstraction which learns a \emph{causal} model that explicitly reasons about which actions and variables affect which variables from collected data, as shown in Fig. \ref{fig:method_comparison} (c). For example, by not depending on door B and the clock, the causal model's predictions about door A are unaffected by those variables and likely to be more accurate than dense models.
Specifically, we prove that each variable's dependence on other variables (or actions) can be determined by a single conditional independence test. Such a test is then carried out by a novel architecture which estimates the conditional mutual information while learning the dynamics model.

Furthermore, by revealing all unnecessary dependencies, certain state variables which no other variables depend on (e.g., the clock) can be omitted for planning, forming a new form of state abstraction. Specifically, our model partitions state variables into those that it can change (\emph{controllable variables}, e.g., doors A and B) with its actions, those that it cannot change but that influence actions' results on those that it can (\emph{action-relevant variables}, e.g., an obstacle that may block door A's motion), and the remainder (\emph{action-irrelevant variables}, e.g., the clock) which have no influence on others and thus can be omitted during planning. Also, in the abstract state, the dynamics model is still free of unnecessary dependencies and exhibits the same generalization benefits. Derived purely from dynamics, our state abstraction includes all controllable variables that the agent can use in the future, enabling it to solve a wider range of tasks than bisimulation which only retains variables specific to a single task.

\textsc{cdl} is compared against state-of-the-art dense models in two simulated environments: a chemical environment with causal relationships of different complexities, and a table-top manipulation environment with challenging rigid body dynamics. We find that \textsc{cdl} learns causal relationships accurately and retains similar prediction accuracy on unseen states to the accuracy on seen states, while the prediction accuracies of dense models drop $60 \sim 90\%$ in some complex environments. When applied to downstream tasks, policies with the proposed causal state abstraction learn with higher sample efficiency and also generalize better than those with dense models.
\section{Related Work}
\label{sec:related}

\subsection{Model-based Reinforcement Learning}

Model-based RL typically involves learning a dynamics model of the environment (including a reward predictor) by maximizing the likelihood of collected trajectories. Then the dynamics model and reward predictor are used for planning \cite{williams2017information, chua2018deep, nagabandi2018neural}, providing synthetic data \cite{kurutach2018model, janner2019trust}, or improving $Q$-value estimates \cite{feinberg2018model, amos2021model}. However, most existing approaches model the dynamics in the dense form, where each state variable at the next time step depends on all current state variables and the action, e.g., as shown in Fig. \ref{fig:method_comparison} (a). This non-causal formulation suffers from spurious correlations and generalizes poorly for out-of-distribution states. Moreover, the reward predictor typically also has the same dense architecture, thus exacerbating the generalization issue.

In dynamics learning, there are models that explicitly consider modularity and sparsity \cite{goyal2019recurrent, goyal2020object, alias2021neural}. However, without modelling the causal relationships, the sparse dependencies learned by those methods could still be spurious correlations and thus lead to poor generalization (see details in appendix Sec. \ref{app:additional_related}).

\subsection{State Abstraction}
\label{subsec:state_abstraction}

Aiming at improving sample efficiency, state abstractions aggregate many states into a single abstract state (e.g., by omitting state variables that are irrelevant to the current task) while preserving policy optimality, thus enabling the agent to learn in a potentially much smaller state space. 
For example, bisimulation \cite{even2003approximate, ravindran2004algebraic, li2009unifying} aggregates states that generate equivalent reward sequences given any action sequence originating from the states. 
Bisimulation is also closely tied to causality: work from \citet{zhang2020invariant} shows that state variables kept by the bisimulation abstraction consist of only the causal ancestors of the reward variable. However, it still learns a dense model for those variables, as shown in Fig. \ref{fig:method_comparison} (b), and is subject to the same generalization issue mentioned earlier.

Furthermore, as the bisimulation abstraction is derived from the reward function, it is often limited to a small range of tasks. Specifically, the abstraction can only generalize to tasks where the new reward function is defined on the same or a subset of causal ancestors. 
For example, in the environment shown in Fig. \ref{fig:method_comparison}, the abstraction learned from opening door A will only include door A, thus is unable to solve tasks like opening door B.
In the real world, one generally wants a robot to be able to learn to perform a wide range of tasks (e.g., manipulating different objects); training this bisimulation abstraction for each task (object) is impractical. In contrast, \textsc{cdl} derives a state abstraction which includes all controllable state variables from the learned dynamics. As a result, it can be applied to a wider range of tasks than bisimulation.

\subsection{Causality in Reinforcement Learning}
\label{subsec:reglarization_based_learning}

Incorporating causality into reinforcement learning has received widespread interest recently, due to its generalization and transferability benefits, in directions that include but are not limited to representation learning \cite{zhang2019learning, zhang2020invariant, sontakke2021causal, volodin2020resolving, tomar2021model, fu2021learning}, policy learning \cite{buesing2018woulda, nair2019causal, mozifian2020intervention, lyle2021resolving, seitzer2021causal, lu2020sample}, and dynamics learning \cite{li2020causal}. Among the literature, the work closest to ours is \citet{wang2021task} which also explicitly learns the causal dependencies but by regularizing the number of variables used when predicting each state variable. However, there is no theoretical guarantee that dynamics models learned by those methods are causal. Moreover, it is found to be difficult for the method from \citet{wang2021task} to balance between prediction performance and regularization. A large regularization penalty often sacrifices the prediction accuracy of rare events whereas a small regularization penalty fails to suppress spurious correlations related to frequent events. In contrast, our method is theoretically sound and does not have a regularization parameter to tune.
\section{Causal Dynamics Learning for Task-Independent State Abstraction (\textsc{cdl})}
\label{sec:approach}
In this section, we introduce \textsc{cdl}, which learns a causal dynamics model and then derives a state abstraction that supports learning a wide range of tasks, from the learned causal relationships.

\subsection{Background}
\subsubsection{Markov Processes}

We consider the agent's interaction with the environment as a Markov Process defined as $\mathcal{M} = (\mathcal{S}, \mathcal{A}, \mathcal{P})$, where $\mathcal{S} \subseteq \mathbb{R}^{d_\mathcal{S}}$ is the $d_\mathcal{S}$-dimensional state space, $\mathcal{A} \subseteq \mathbb{R}^{d_\mathcal{A}}$ is the $d_\mathcal{A}$-dimensional action space, $\mathcal{P}$ is the transition function. The state space consists of $d_\mathcal{S}$ state variables, one for each dimension of the state space, denoted as $\{\mathcal{S}^i\}_{i=1}^{d_\mathcal{S}}$.

\subsubsection{Causal Graphical Models}
\label{subsec:causal_graphical_model}
A causal graphical model \cite{pearl2009causality} is defined by a distribution $p_X$ over $d$ random variables $X^1, \dots, X^d$ and a Directed Acyclic Graph $\mathcal{G}=(V, E)$. Each node $i \in V = \{1, \dots, d\}$ is associated with a random variable $X^i$ and each edge $(i \rightarrow j) \in E$ represents that $X^i$ is a \textbf{direct cause} of $X^j$, i.e., $X^j$ depends on $X^i$ during the data generation process. 
Then the distribution can be specified as
\begin{equation}
    p_X(x^1, \dots, x^d) = \prod_{i=1}^d p_{X^i}(x^i | \textbf{PA}_i),
    \label{eq:causal_graphical_model}
\end{equation}
where $\textbf{PA}_i$ is the set of parents of the node $i$ in the graph $\mathcal{G}$.

For simplicity, in the remainder of the paper, we denote $v^{1:n}$ (or $v_{1:n}$) for an abbreviation of a set of variables $\{v^i\}_{i=1}^n$ (or $\{v_i\}_{i=1}^n$) and will omit the subscript of distributions, i.e., $p_V(v) = p(v)$ and $p_V(v | w) = p(v | w)$.

\begin{figure}
  \centering
  \includegraphics[width=0.7\columnwidth]{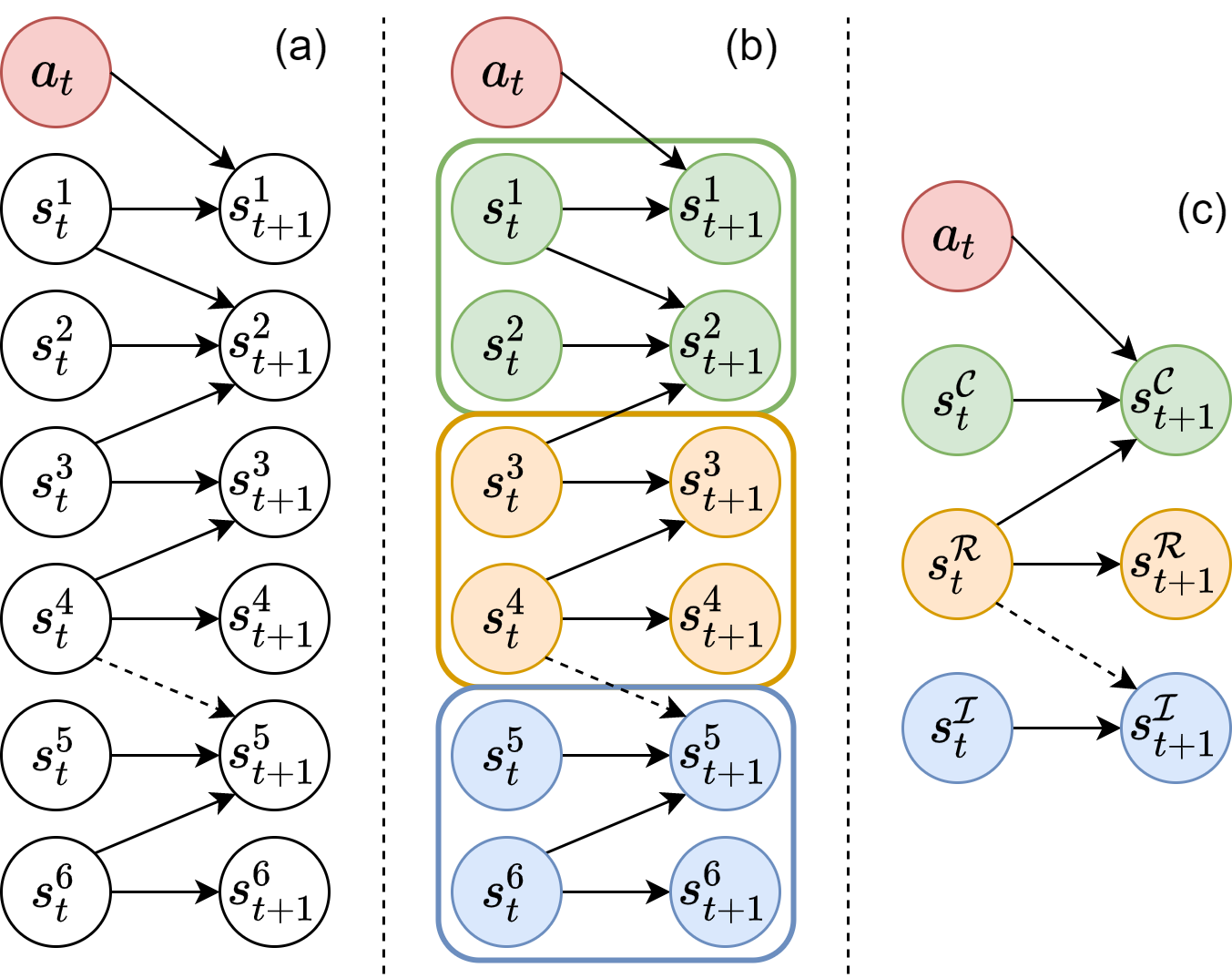}
  \vspace{-10pt}
  \caption{\small \textbf{(a)} an example causal dynamics model. \textbf{(b)} state variables can be split into three types: controllable (green), action-relevant (orange), and action-irrelevant (blue). The dashed arrow represents whether it exists does not affect $s^5$ to be action-irrelevant. \textbf{(c)} the causal graph can be split into three subgraphs, one for each type of state variable.}
  \label{fig:causal_illustration}
  \vspace{-10pt}
\end{figure}

\subsection{Problem Definition}
\label{subsec:problem_definition}

We begin with a formal definition of controllable, action-relevant, and action-irrelevant state variables. If we are given the causal graphical model of a Markov Process with $V=\{s^{1 : \dS}_t, a_t, s^{1 : \dS}_{t+1} \}$ as nodes and $E$ as edges describing causal relationships from $s^{1 : \dS}_t$ and $a_t$ to $s^{1 : \dS}_{t+1}$, ancestors of a state variable $s^i$ are defined as all nodes that have a directed path leading to node $s^i$ (not necessarily from the immediate previous time step but can be from any previous step). For example, for the causal dynamics model shown in Fig. \ref{fig:causal_illustration} (a), $s^4$ is an ancestor of $s^2$ as there is a path of $s^4_t \rightarrow s^3_{t+1} \rightarrow s^2_{t+2}$. Descendants of nodes are defined in the same way but in the opposite direction. Then we have:

\textbf{Definition 1} (Controllable State Variables) $s^\mathcal{C}$ are the descendants of the action $a_t$.

\textbf{Definition 2} (Action-Relevant State Variables) $s^\mathcal{R}$ are ancestors of controllable state variables, excluding those already belonging to $s^\mathcal{C}$.

\textbf{Definition 3} (Action-Irrelevant State Variables) $s^\mathcal{I}$ are those that belong to neither $s^\mathcal{C}$ nor $s^\mathcal{R}$.

In the above definitions, $\mathcal{C}, \mathcal{R}$, and $\mathcal{I}$ are the set of state dimension indices for controllable, action-relevant, and action-irrelevant state variables respectively.

Given these definitions, the type of each state variable in the example causal dynamics model is shown in Fig. \ref{fig:causal_illustration} (b). Further in (c), one may notice that the causal graph can be split into three parts, allowing us to rewrite the transition probabilities as $\mathcal{P}(s_{t+1} | s_{t}, a_t)= p(s^\mathcal{C}_{t+1} | s^\mathcal{C}_{t}, s^\mathcal{R}_{t}, a_t) \cdot p(s^\mathcal{R}_{t+1} | s^\mathcal{R}_{t}) \cdot p(s^\mathcal{I}_{t+1} | s^\mathcal{R}_{t}, s^\mathcal{I}_{t})$,
where the edge from $s^\mathcal{R}_{t}$ to $s^\mathcal{I}_{t+1}$ is optional as it does not change the split of state variables ($s^\mathcal{I}$ are still neither $a_t$'s descendants nor $s^\mathcal{C}$'s ancestors).

Then \textsc{cdl} forms the state abstraction $\phi$ by omitting action-irrelevant state variables, i.e., $\phi(s_t) = (s^\mathcal{C}_t, s^\mathcal{R}_t)$, and the dynamics in the abstract space can be expressed by removing the subgraph involving action-irrelevant state variables, remaining a causal dynamics model itself, as follows:
\begin{align}
    \mathcal{P}(\phi(s_{t+1}) | \phi(s_t), a_t) = p(s^\mathcal{C}_{t+1} | s^\mathcal{C}_{t}, s^\mathcal{R}_{t}, a_t) \cdot p(s^\mathcal{R}_{t+1} | s^\mathcal{R}_{t}). \nonumber
\end{align}

This state abstraction $\phi$ can be used to solve any \textit{actively-accomplishable} downstream task.
Here, downstream tasks are those defined in the same Markov Process so that the agent can use $\phi$ to solve the task by learning the provided rewards.
Meanwhile, actively-accomplishable means that the reward function of the task only depends on controllable and action-related state variables $(s^\mathcal{C}, s^\mathcal{R})$.
Additionally, if the task involves extra variables $\mathcal{V}$, e.g., a \textit{varying} goal $g_t$ for certain controllable state variables to reach, those variables should also be provided so that the agent can learn the reward accurately.
Notice that actively-accomplishable tasks do not cover those involving action-irrelevant state variables (e.g., for the example in Fig. \ref{fig:method_comparison}, a reward of $+1$ for opening door A when the clock shows 1 pm and $0$ otherwise).
However, as the reward function of a task can be arbitrarily designed using any state variable, being able to solve all tasks means that no state variable can be omitted (i.e., no state abstraction). We assume that in practice, it is relatively uncommon for a task’s reward function to involve action-irrelevant state variables, making $\phi$ fairly generally applicable.

So far, we have defined three types of state variables and the derived state abstraction for a known causal dynamics model. However, for real-world problems, such a model is usually not accessible. Instead, agents can only collect transition data via its interactions with the environment. Hence, this paper introduces a novel method that: (1) learns a \textbf{causal} dynamics model $F_\theta: \mathcal{S} \times \mathcal{A} \rightarrow \mathcal{S}$, (2) learns a transition collection policy $\pi: \mathcal{S} \rightarrow \mathcal{A}$ to learn $F_\theta$ efficiently, (3) derives the state abstraction $\phi: \mathcal{S} \rightarrow \mathcal{S}^\mathcal{C} \times \mathcal{S}^\mathcal{R}$ and dynamics $F^\phi_\theta$ in the abstract space, (4) learns a reward predictor for any actively-accomplishable task in the abstract space $R_\varphi: \phi(\mathcal{S}) \times \mathcal{A} \ (\times \mathcal{V}) \rightarrow \mathbb{R}$, and (5) uses planning methods to solve the task with the learned $F^\phi_\theta$ and $R_\varphi$.

\subsection{Causal Dynamics Learning}

The key challenge when learning a causal dynamics model is to determine whether a causal edge exists between two state variables, i.e., $s_t^i \rightarrow s^j_{t+1}$. First, adapting the work from \citet{mastakouri2021necessary}, we present a method for inferring whether such a causal relationship exists, based on the following assumptions about the ground truth dynamics model:

\textbf{Assumptions}\\
A1. Causal Markov condition and Faithfulness in the underlying dynamics (definitions in appendix Sec. \ref{app:definitions}).\\
A2. The state is fully observable and the dynamics is Markovian.\\
A3. The edge $s_t^i \rightarrow s^i_{t+1}$ exists for all state variables $s^i$.\\
A4. No simultaneous or backward edges in time, i.e., for all $i, j$, $s_t^i \nrightarrow s^j_{t}$ and $s_t^i \nrightarrow s^j_{t-1}$.\\
A5. The transitions for each state variable are independent, i.e., $\mathcal{P}(s_{t+1}|s_t, a_t) = \prod_{j=1}^\dS p(s^j_{t+1}|s_t, a_t)$

Except for A4, which requires no redundant information in state variables (e.g., state variables that include both joint angles and the end-effector of a robot arm), and A5, which does not necessarily hold for rich observation spaces (e.g., images), these assumptions are commonly made for causal inference and dynamic systems. Moreover, for partially observable or high-dimensional state spaces, low-dimensional disentangled representations that encode the space can be learned to adhere to A2 and A5.

\begin{theorem}
\label{thm:causal_sufficient_condition}
Assuming A1 - A5, we define the conditioning set $\{a_t, s_t \setminus s^i_t\}=\{a_t, s^1_t, \dots, s^{i-1}_t, s^{i+1}_t, \dots, s^\dS_t\}$. Then, for any two state variables at indices $i$ and $j$, if $s_t^i \notind s^j_{t+1} | \{a_t, s_t \setminus s^i_t\}$, then $s_t^i \rightarrow s^j_{t+1}$. Similarly, if $a_t \notind s^j_{t+1} | s_t$, then $a_t \rightarrow s^j_{t+1}$.
\end{theorem}

Proof in appendix Sec. \ref{app:theorem_proof}. This result shows that the causal relationship between state variables (or between the action and any state variable) can be inferred with one Conditional Independence Test (\textsc{cit}). For simplicity, in the remainder of the paper, we will only describe the \textsc{cit} between two state variables, $s_t^i \notind s^j_{t+1} | \{a_t, s_t \setminus s^i_t\}$. For the test between the action and the state variable, $a_t \notind s^j_{t+1} | s_t$, the same method applies by changing $s_t^i$ to $a_t$ and the conditioning set according to Theorem \ref{thm:causal_sufficient_condition}. 

In theory, the \textsc{cit} between $s_t^i$ and $s^j_{t+1}$ can be made by measuring the Conditional Mutual Information, $\textsc{cmi}^{ij}$:
\begin{align}
\label{eq:cmi}
& \mathop{\mathbb{E}}\limits_{s_t,a_t,s^j_{t+1}}\left[\log\frac{p(s^i_t, s^j_{t+1} | \{a_t, s_t \setminus s^i_t\})}{p(s^i_t | \{a_t, s_t \setminus s^i_t\})p(s^j_{t+1} | \{a_t, s_t \setminus s^i_t\})}\right] \nonumber\\
=& \mathop{\mathbb{E}}\limits_{s_t,a_t,s^j_{t+1}}\left[\log\frac{p(s^j_{t+1} | a_t, s_t)}{p(s^j_{t+1} | \{a_t, s_t \setminus s^i_t\})}\right],
\end{align}
where the expectation is over the joint distribution of $\{s_t,a_t,s^j_{t+1}\}$, all $p$ are the ground truth conditional densities, and the derivation of Eq. \eqref{eq:cmi} is in the appendix Sec. \ref{app:cmi_eq_derivation}. If $\textsc{cmi}^{ij} \ge \epsilon$ where $\epsilon$ is a pre-defined threshold, it suggests that $s^i_t$ is necessary to predict $s^j_{t+1}$ accurately, $s_t^i \notind s^j_{t+1} | \{a_t, s_t \setminus s^i_t\}$ is true, and the causal edge $s_t^i \rightarrow s^j_{t+1}$ exists.

In practice, as the ground truth joint distribution and conditional densities are unknown, the expectation is computed over the collected transition data $\mathcal{D}$ and \textsc{cdl} learns predictive models $\hat{p}(s^j_{t+1} | a_t, s_t), \hat{p}(s^j_{t+1} | \{a_t, s_t \setminus s^i_t\})$ to approximate conditional densities.

\begin{figure}
  \centering
  \includegraphics[width=0.9\columnwidth]{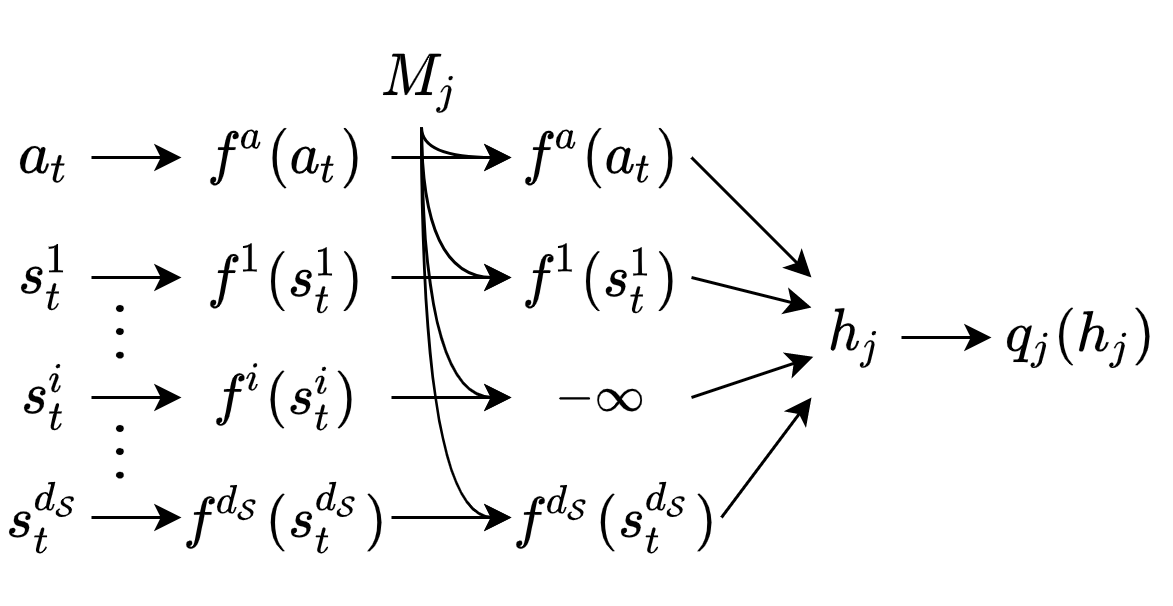}
  \vspace{-15pt}
  \caption{\small The predictive model for the state variable $s^j_{t+1}$. Different conditional densities can be represented by applying different masks $M_j$. $\hat{p}(s^j_{t+1} | \{a_t, s_t \setminus s^i_t\})$ is shown as an example in the figure.}
  \label{fig:architecture}
  \vspace{-15pt}
\end{figure}

However, computing $\textsc{cmi}^{ij}$ for all state variable pairs requires training $(d_\mathcal{S})^2$ predictive models, which is intractable. Instead, we develop a novel architecture and training paradigm which reduces the requirement to $\dS$ models. 
As shown in Fig. \ref{fig:architecture}, to predict each state variable $s^j_{t+1}$:
(1) the action and all current state variables are individually mapped to features $f^a_j(a_t), f^1_j(s^1_t), \dots, f^\dS_j(s^\dS_t)$ where each feature is a $d_f$-dimensional vector;
(2) certain features are masked to $-\infty$ according to a binary map $M_j$; 
(3) the overall feature $h_j$ is computed by taking the element-wise max of all features; and
(4) a predictive network $q_j$ takes $h_j$ as input and predicts the distribution of $s^j_{t+1}$. 
This architecture can represent conditional densities with different masks $M_j$:
(1) for $\hat{p}(s^j_{t+1} | a_t, s_t)$, none of the features is masked; 
(2) for $\hat{p}(s^j_{t+1} | \{a_t, s_t \setminus s^i_t\})$, the feature $f^i_j(s^i_t)$ is masked to $-\infty$ so that it will not be used to predict $s^j_{t+1}$, as shown in Fig. \ref{fig:architecture}; and 
(3) for $\hat{p}(s^j_{t+1} | \textbf{PA}_{s^j})$, after deriving parents of $s^j_{t+1}$ from $\{\textsc{cmi}^{ij}\}_{i=1}^{\dS}$), all non-parent features are masked to $-\infty$.

The architecture described above predicts one state variable $s^j_{t+1}$, and the whole causal dynamics model $F_\theta$ consists of $\dS$ such models, where $\theta$ parameterizes all feature extractors $\{f^a_j, f^{1:\dS}_j\}_{j=1}^\dS$ and predictive networks $q_{1:\dS}$.
To train $\theta$, we maximize the following log-likelihood:
\begin{align}
\label{eq:dynamics_loss}
\mathcal{L}_\theta = \sum\limits_{j=1}^\dS \big[& \log\hat{p}(s^j_{t+1} | a_{t}, s_t) + \log\hat{p}(s^j_{t+1} | \{a_{t}, s_t \setminus s^i_t\}) \nonumber \\
& + \log\hat{p}(s^j_{t+1} | \textbf{PA}_{s^j})\big],
\end{align}
where $i$ is uniformly sampled from $\{1,\dots,\dS\}$ for each $j$, and $\textbf{PA}_{s^j}$ are inferred from $\{\textsc{cmi}^{ij}\}_{i=1}^{\dS}$ learned so far. In Equation \ref{eq:dynamics_loss}, the first two predictive likelihoods train models necessary to evaluate $\textsc{cmi}^{ij}$, and the last prediction likelihood finetunes the performance of the inferred causal dynamics model $\hat{p}(s^j_{t+1} | \textbf{PA}_{s^j})$. We split the collected transition data $\mathcal{D}$ into the training part used to maximize $\mathcal{L}_\theta$ and the validation part for evaluating $\textsc{cmi}$.

After training, \textsc{cdl} evaluates the causal graph by checking if each $\textsc{cmi}^{ij} \ge \epsilon$ and derives the learned state abstraction $\phi(s) = (\hat{s}^\mathcal{C}, \hat{s}^\mathcal{R})$ from the learned causal graph, according to Definitions 1-3 (for a ground truth variable or distribution $x$, we denote $\hat{x}$ as its learned prediction or distribution). The dynamics in the abstract space $F^\phi_\theta$ can also be derived by omitting the prediction networks for action-irrelevant state variables $\hat{s}^\mathcal{I}$.

\subsection{Policy Learning for Transition Collection}

The collected transition data $\mathcal{D}$ are important to accurate causal dynamics learning, as they are used in predictive model training (Eq. \ref{eq:dynamics_loss}) and $\textsc{cmi}$ evaluations (Eq. \ref{eq:cmi}). An ideal transition collection policy $\pi$ would cover all state-action pairs to expose causal relationships thoroughly and actively explore states where the causal dynamics model is not accurate.

To this end, an RL agent is used for transition collection with a reward function that is the prediction difference between the dense predictor and the causal predictor learned so far:
\begin{align}
\label{eq:policy_reward}
r_t = \tanh \left(\tau \cdot \sum\limits_{j=1}^\dS \log \frac{\hat{p}(s^j_{t+1} | s_t, a_t)}{\hat{p}(s^j_{t+1} | \textbf{PA}_{s^j})}\right),
\end{align}
where $\tau$ is the scaling factor and $\tanh$ is used to keep the reward bounded. This reward motivates taking transitions where the dense predictor is better than the causal predictor, which usually suggests the learned causal graph is inaccurate. 

\subsection{Policy Learning for Downstream Tasks}
\label{subsec:task_learning}

When solving actively-accomplishable downstream tasks, like many MBRL algorithms, \textsc{cdl} simultaneously (1) learns a reward predictor with the abstract state, action, and any provided extra variables as input, $R_\varphi: \phi(\mathcal{S}) \times \mathcal{A} \ (\times \mathcal{V}) \rightarrow \mathbb{R}$, and (2) uses a planning algorithm with the learned dynamics and reward predictor for action selection. As the reward predictor is learned in an abstract space rather than the full state space, it is more sample-efficient and less vulnerable to spurious correlations brought about by excessive information (i.e., action-irrelevant state variables). Meanwhile, planning in the abstract space also reduces the computation cost by relieving the need to roll out action-irrelevant state variables.

The reward predictor is modeled as a neural network and trained by minimizing the prediction error,
\begin{equation}
    \varphi^* = \argmin_\varphi \mathbb{E}_{(s_t, a_t, g_t, r_t) \sim \mathcal{B}} \ \mathcal{L}(R_\varphi(\phi(s_t), a_t, g_t), r_t),
    \label{eq:reward_loss}
\end{equation}
where $\mathcal{B}$ is the task data collected so far, and $\mathcal{L}$ can take any loss function (we experiment with the absolute value of the prediction error).

For planning, we use the cross entropy method (\textsc{cem}, \citet{rubinstein1997optimization}), a population-based optimization algorithm, to search for the best action with the learned dynamics and reward predictor. For each time step $t$, depending on whether the action is continuous or discrete, \textsc{cem} initializes a time-dependent belief over the optimal action sequence $a_{t:t+L} \sim \mathcal{N}(\mu_{t:t+L}, \sigma^2_{t:t+L})$ or $\text{Categorical}(\alpha_{t:t+L})$ where $L$ is the planning length. Starting from a unit normal distribution for $\mathcal{N}$ or a uniform distribution for $\text{Categorical}$, it repeatedly samples $J$ candidate action sequences, evaluates them based on cumulative rewards, and updates the action belief to the distribution of the top $K$ candidates. After $N$ iterations, the planner returns $\mu_t$ or $\argmax \alpha_t$ as the current optimal action.
\section{Experiments}
\label{sec:experiments}

Our experiments examine the following hypotheses:\\
1. \textsc{cdl} can infer the causal graph, with less tuning and better accuracy than a regularization-based method \cite{wang2021task} (Sec. \ref{subsubsec:causal_graph_accuracy_results}).\\
2. Causal models learned by \textsc{cdl} generalize better than dense models in both dynamics and downstream task learning on unseen states (Sec. \ref{subsubsec:prediction_results}).\\
3. The transition collection policy learned by \textsc{cdl} improves the accuracy of causal dynamics learning compared to a uniform policy or the policy learned from the curiosity reward \cite{pathak2017curiosity}. (Sec. \ref{subsec:data_collection_policy_results}).\\
4. State abstractions derived by \textsc{cdl} improve sample-efficiency when learning downstream tasks compared to dense models which use the full state space (Sec. \ref{subsec:task_results}).\\
5. Policies learned by \textsc{cdl} for downstream tasks generalize better than those learned by dense models (Sec. \ref{subsec:task_results}).

\subsection{Environments}
\label{subsec:environment}

\begin{figure}
  \centering
  \includegraphics[width=0.9\columnwidth]{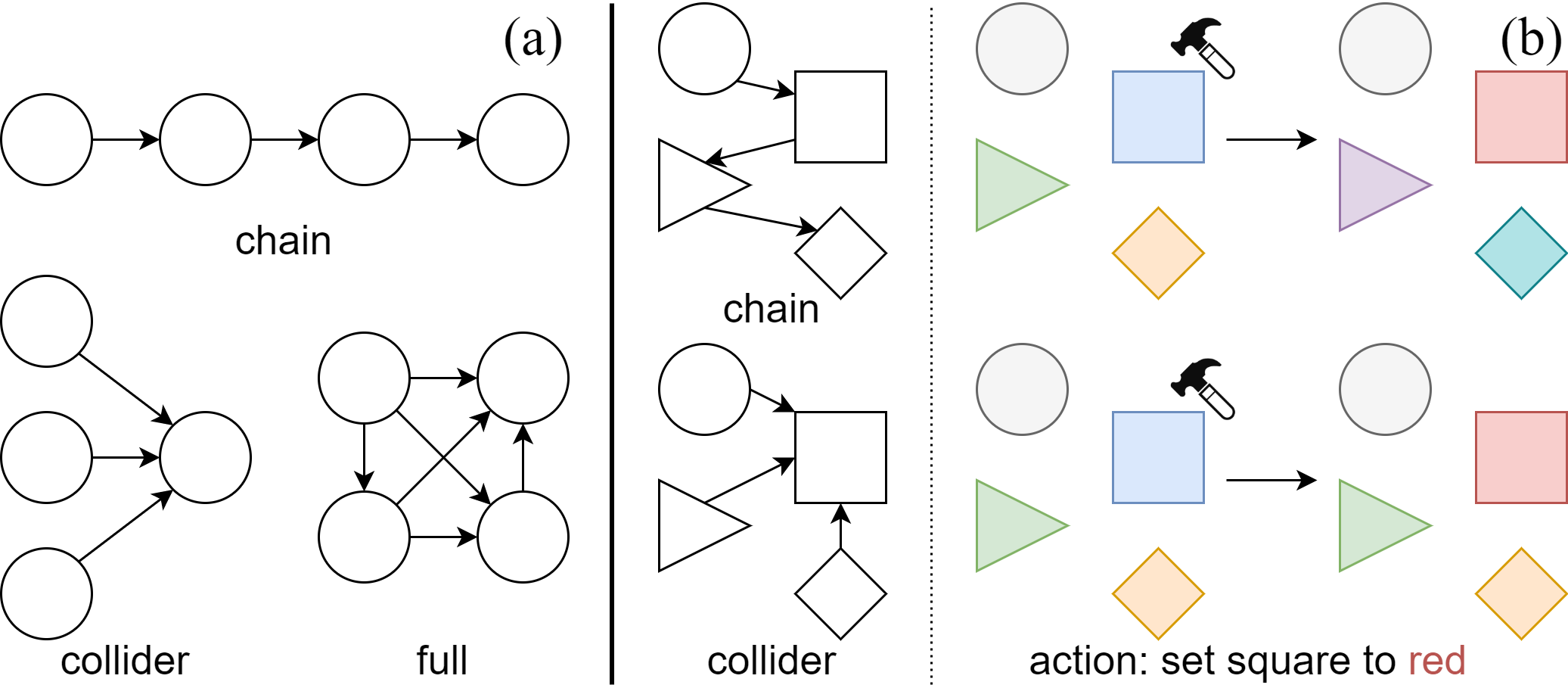}
  \vspace{-10pt}
  \caption{\small (a) different types of causal graphs. (b) illustration of the chemical environment (left: ground truth causal graphs, right: transitions after applying the action).}
  \label{fig:chemical_env}
  \vspace{-10pt}
\end{figure}

\begin{figure}
  \centering
  \includegraphics[width=0.9\columnwidth]{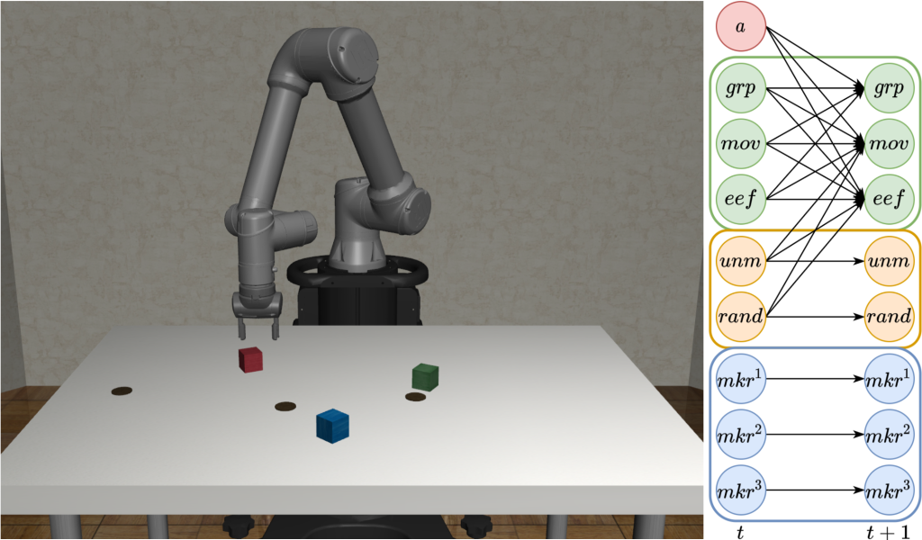}
  \vspace{-10pt}
  \caption{\small \textbf{(Left)} the manipulation environment. \textbf{(Right)} the learned causal graph in the object level which recovers the ground truth relationship between the state variables and the action.}
  \label{fig:robosuite}
  \vspace{-15pt}
\end{figure}

We use (1) the Chemical environment modified from \citet{ke2021systematic} (Fig. \ref{fig:chemical_env} (b)) to examine \textsc{cdl}'s performance on causal graphs with different complexities, and (2) a manipulation environment (Fig. \ref{fig:robosuite}) to validate \textsc{cdl} with complex rigid-body dynamics.

In the chemical environment, there are 10 objects whose x,y positions are continuous and whose colors are one of 5 colors; the action can change a chosen object to a specified new color. The interactions between objects take place according to the underlying causal graph - when one object is set to a new color, all its descendants change their colors in the order of the graph following a predefined conditional probability table. The dynamics of colors and positions is independent for all objects: positions are sampled from a normal distribution every step. We consider three graph structures shown in Fig. \ref{fig:chemical_env} (a): full, chain, and collider.

In the manipulation environment implemented with the \texttt{robosuite} simulation framework \cite{robosuite2020}, there are three objects that the robot can interact with: a \textit{movable} red cube that the robot can manipulate (denoted as $mov$), an \textit{unmovable} green cube that is fixed on the table ($unm$), and a \textit{randomly moving} blue cube that also cannot be moved ($rand$). The purpose of the randomly moving object is to test whether \textsc{cdl} can learn that its motion is not caused by the robot's actions. There are also three randomly moving non-interactable markers ($mkr^1$, $mkr^2$, and $mkr^3$) on the table, as distractors to introduce potential spurious correlations. The state space consists of the robot end-effector (\textsc{eef}) location ($\mathbb{R}^3$), gripper ($grp$) joint angles ($\mathbb{R}^2$), and locations of objects and markers ($6 \times \mathbb{R}^3$). The action space includes \textsc{eef} location displacement ($\mathbb{R}^3$) and the degree to which the gripper is opened ($[0, 1]$). In each episode, the objects and markers are reset to randomly sampled poses on the table.

\subsection{Implementation}

\subsubsection{baselines}

We compare \textsc{cdl} with the following baselines:\\
1. \textbf{Reg}: regularization-based causal dynamics learning \cite{wang2021task}, where the causal mask $M$ is learned as trainable parameters rather than inferred from \textsc{cmi}.\\
2. \textbf{Modular}: using a separate network to predict each state variable \cite{ke2021systematic}, which can be represented by our method when none of the features are masked.\\
3. \textbf{GNN}: a graph neural network, following \cite{ke2021systematic}, we consider the C-SWM model \cite{kipf2019contrastive} which assumes dense dependence for each state variable.\\
4. \textbf{Monolithic}: a multi-layer perceptron (MLP) network that takes all state variables and actions as inputs and predicts the next step values of all state variables, which is commonly used in \textsc{mbrl}.

For a fair comparison, we use the same architecture for \textsc{cdl}, Reg, and Modular. For GNN and Monolithic whose architectures are different from the others, we try our best to give them the same number of parameters as others. See architectures of all methods in appendix Sec. \ref{app:architectures}.

\subsubsection{Causal Dynamics Learning}

We use MLP networks for feature extractors $f^a_{1:\dS}, f^{ 1:\dS}_{1:\dS}$ and predictive networks $q_{1:\dS}$. For continuous state variable $s^j_{t+1}$, $q_j$ outputs the mean and standard deviation of a normal distribution. For discrete state variable, $q_j$ outputs the logits of a categorical distribution. The threshold to infer causal relationships from \textsc{cmi} varies across environments.  We specify it along with other hyperparameters in appendix Sec. \ref{app:dynamics_learning}. Empirically, we find that reducing the frequency of optimizing causal prediction likelihood (i.e., $\log\hat{p}(s^j_{t+1} | \textbf{PA}_{s^j})$ in Eq. \ref{eq:dynamics_loss}) can stabilize training, so we only include this term once every 10 updates of the dynamics parameter $\theta$ according to Eq. \ref{eq:dynamics_loss} .

\subsubsection{Transition Collection Policy}

For the simple chemical environment, we use a random policy to collect transitions as it is enough to cover most of the state-action pairs. For the manipulation environment, we use Proximal Policy Optimization (PPO, \citet{schulman2017proximal}) as our RL agent. Meanwhile, as exploring the whole state space for manipulation domains is still a challenging research topic and not the focus of this paper, to reduce the complexity of exploration, we follow \citet{nasiriany2021maple} and modify PPO's action space to long-horizon skills, such as position reaching and object grasping/pushing (details in appendix Sec. \ref{app:transition_collection_policy_implementation}). Notice that dynamics learning still uses low-level raw motor actions described in Sec. \ref{subsec:environment}.

\subsection{Causal Dynamics Learning Evaluation}
\label{subsec:dynamics_results}

After training all methods on the same set of collected transition data (details in appendix Sec. \ref{app:transition_collection_policy_implementation}), we evaluate the quality of their learned dynamics in the following aspects:

\textit{Causal Relationship Inference}. We compare the causal graph inferred by \textsc{cdl} and Reg with the ground truth graph, in terms of edge accuracy.\\
\textit{Predicting Future States}. Given the current state and a sequence of actions, we evaluate the accuracy of each method's $k$-step prediction, for states both in and out of the training distribution (denoted as ID and OOD states). For the chemical environment, we evaluate color prediction in terms of mean accuracy (MA). For manipulation, log-likelihoods is used.\\
We also provide results using other metrics in the appendix Sec. \ref{app:dynamics_results}.

In the remainder of the paper, the performances shown for each method are mean and standard deviation computed from 3 models with different seeds.

\subsubsection{Causal Relationship Inference}
\label{subsubsec:causal_graph_accuracy_results}

\begin{table}
\centering
\caption{Causal Graph Accuracy (in $\%$) for \textsc{cdl} and Reg}
\vspace{-0pt}
\begin{small}
\begin{tabular}{ccc}
\toprule
\textbf{Environment} & \textbf{\textsc{cdl}} (Ours) & \textbf{Reg} \\
\midrule
Chemical (Collider) & \textbf{100.0} $\pm$ 0.0  & 99.4 $\pm$ 0.4 \\
Chemical (Chain)    & \textbf{100.0} $\pm$ 0.1  & 99.7 $\pm$ 0.1 \\
Chemical (Full)     & \textbf{99.1} $\pm$ 0.1   & 97.7 $\pm$ 0.4 \\
Manipulation        & \textbf{90.2} $\pm$ 0.3   & 84.4 $\pm$ 0.5 \\
\bottomrule
\end{tabular}
\end{small}
\vspace{-10pt}
\label{tab:causal_graph_accuracy}
\end{table}

As shown in Table. \ref{tab:causal_graph_accuracy}, in all environments, \textsc{cdl} learns the causal graph more accurately than Reg. Other baselines are not compared as they do not learn causal relationships. Fig. \ref{fig:robosuite} right shows the causal graph learned by \textsc{cdl} for the manipulation environment in the object level, and other quantitative and qualitative results can be found in appendix Sec. \ref{app:causal_graph_results} which collectively indicate that  \textsc{cdl} learns both fewer false positives (i.e., spurious correlations) and false negatives (i.e., missing necessary dependencies).

\subsubsection{Predicting Future States}
\label{subsubsec:prediction_results}

\begin{figure}
  \centering
  \includegraphics[width=1.0\columnwidth]{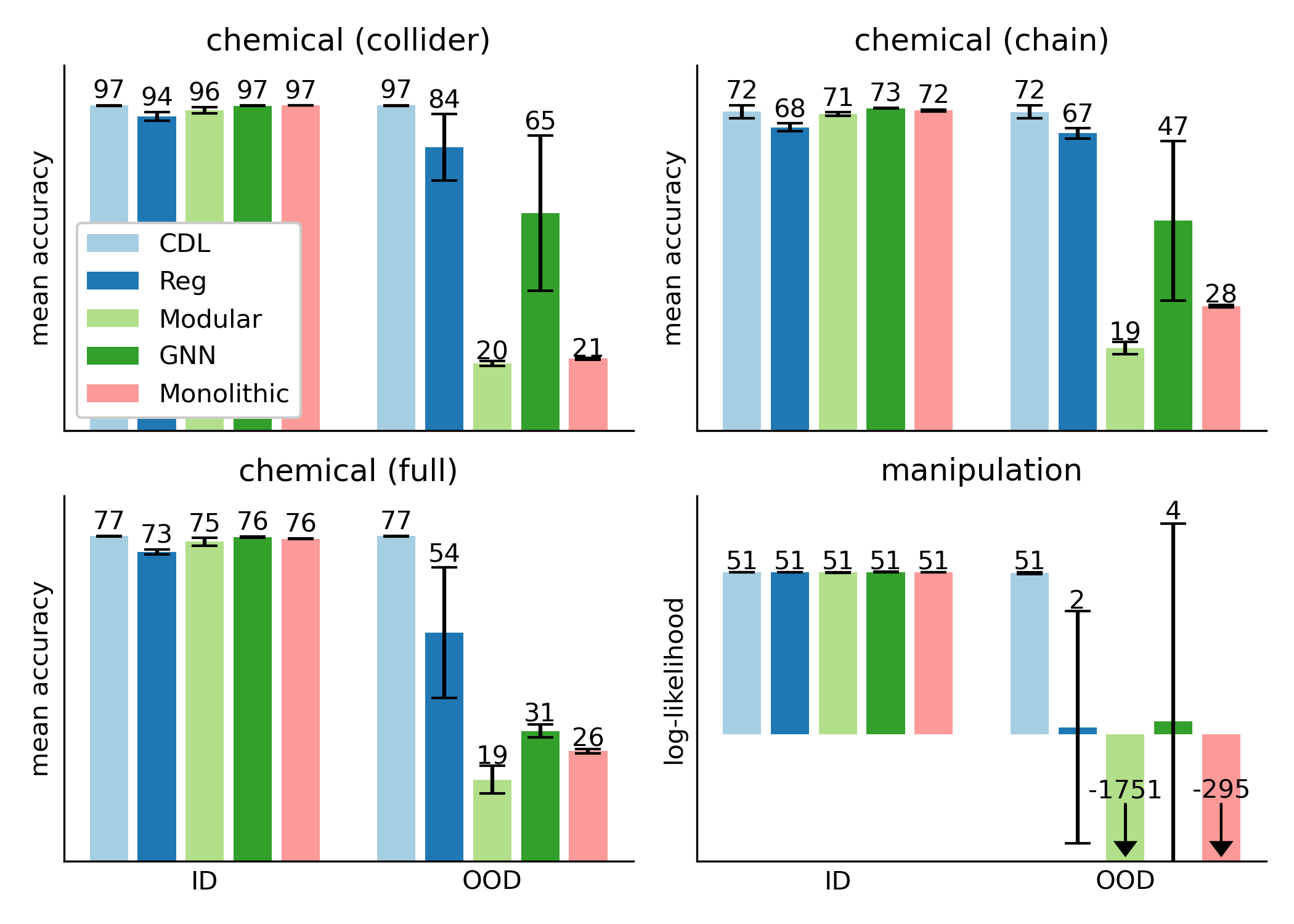}
  \vspace{-25pt}
  \caption{\small 1-step prediction performance on ID and OOD states. The mean is marked on top of each bar.}
  \label{fig:prediction_results}
  \vspace{-15pt}
\end{figure}

We evaluate each method for $1 \sim 5$-step prediction on 5K transitions, for both ID and OOD states. To create OOD states, we change object positions in the chemical environment and marker positions in the manipulation environment to unseen values, and the prediction is measured on other unchanged state variables. The 1-step prediction performance for each method is shown in Fig. \ref{fig:prediction_results} (the Modular and Monolithic in the bottom right plot are cropped due to their bad performances, and their means are labeled instead), and results for $2 \sim 5$-steps are in appendix Sec. \ref{app:prediction_results}. For ID states, all methods show similar prediction performance. However, for OOD states, the performance of dense models drops significantly as they suffer from spurious correlations while \textsc{cdl} and Reg mask those out with the learned dependencies. However, Reg also performs badly in the manipulation domain, because the trade-off between prediction accuracy and regularization leads to an inaccurate causal graph.

\subsection{Transition Collection Policy Learning Evaluation}
\label{subsec:data_collection_policy_results}

\begin{table}
\centering
\begin{small}
\caption{Causal Graph Accuracy for \textsc{cdl} with Different Transition Collection Policies}
\begin{tabular}{cccc}
\toprule
\textbf{Method}             & \textbf{PredDiff} (Ours) & \textbf{Uniform} & \textbf{Curiosity} \\
\midrule
\textbf{Accuracy} $(\%)$    & \textbf{90.2} $\pm$ 0.3 & 89.1 $\pm$ 0.6 & 88.9 $\pm$ 0.2 \\
\bottomrule
\end{tabular}
\end{small}
\vspace{-20pt}
\label{tab:policy_results}
\end{table}

To evaluate the benefits of the proposed transition collection policy (denoted as \textbf{PredDiff}), we compare its performance with the following methods, in terms of causal graph accuracy in the manipulation environment, \\
\textbf{Uniform}: a fixed policy that uniformly selects between skills and skill parameters instead of active exploration;\\
\textbf{Curiosity}: a policy that uses the prediction error of the causal predictor $\hat{p}(s^j_{t+1} | \textbf{PA}_{s^j})$ as rewards which encourages transitions where prediction errors are high \cite{pathak2017curiosity};

Table \ref{tab:policy_results} shows the accuracies of the causal graphs learned by \textsc{cdl} from the same amount of data (1.8M transitions) but collected by those three different policies, measured in the same way as Sec. \ref{subsubsec:causal_graph_accuracy_results}. Our PredDiff policy learns the graph with the highest accuracy. Though the accuracy differences between methods seem small, in a causal graph representing complex rigid body dynamics with $23 \times 24$ potential dependencies, $1\%$ accuracy difference means prediction errors for 5 $\sim$ 6 edges, which may severely harm the generalizability of the learned causal dynamics model. Furthermore, as shown in appendix Sec. \ref{app:data_collection_policy_results}, causal graphs learned by Uniform have more spurious correlations as it doesn't actively explore transitions where the causal graph is not learned well, and the Curiosity policy suffers from procrastination (i.e., repeats the transition with large stochasticity to maximize rewards) and fails to expose other causal dependencies. 

\subsection{Downstream Tasks Learning Evaluation}
\label{subsec:task_results}

We evaluate each method with the following downstream tasks in the chemical environment (C) and in the manipulation environment (M): \\
\textbf{Match} (C): match the object colors with goal colors individually, \\
\textbf{Reach} (M): move the end-effector to the goal position, \\
\textbf{Lift} (M): lift the movable object to the goal position,\\
\textbf{Stack} (M): stack the movable object on the top of the unmovable object.

The tasks' reward functions are unknown to the agent and are specified in appendix Sec. \ref{app:task_results}. During training, the dynamics model is frozen and only the reward predictor is learned as described in Sec. \ref{subsec:task_learning}.

\begin{figure}
  \centering
  \includegraphics[width=1.0\columnwidth]{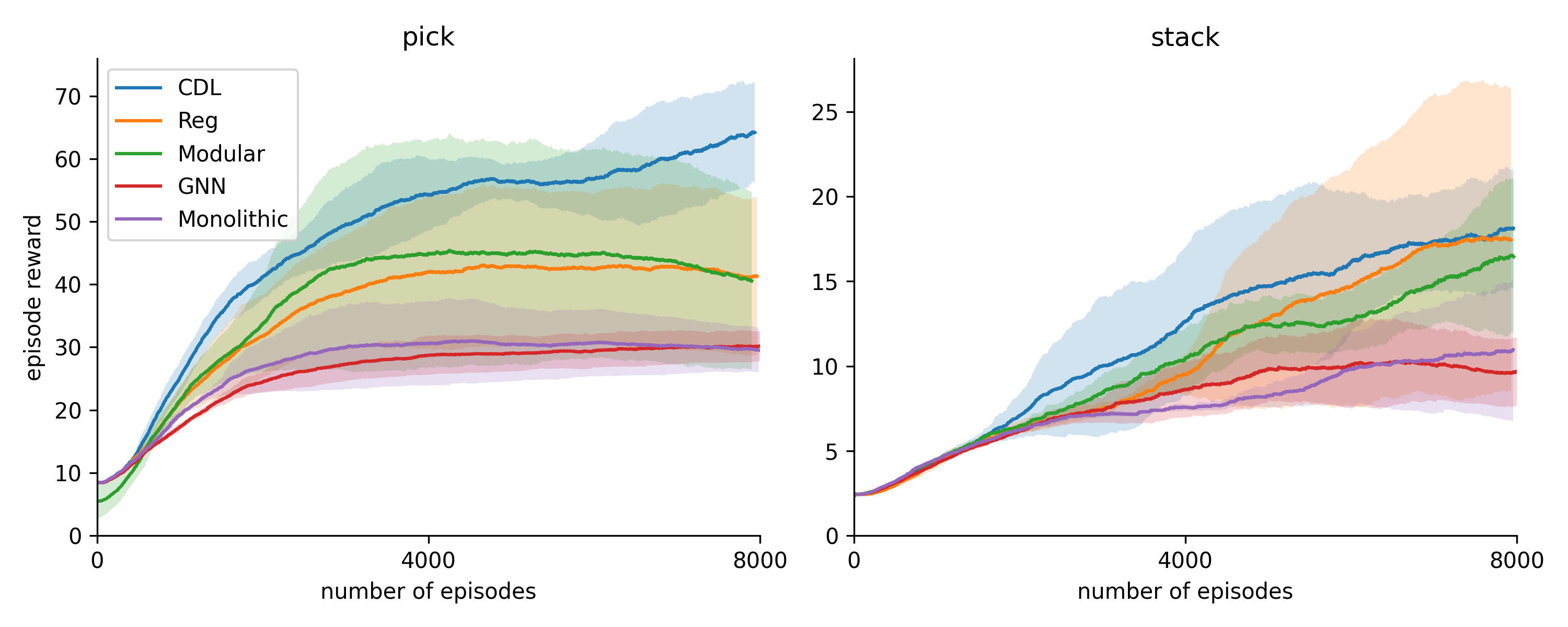}
  \vspace{-25pt}
  \caption{\small Training curves for two downstream tasks.}
  \label{fig:learning_curve}
  \vspace{-10pt}
\end{figure}

Compared to dense baselines, \textsc{cdl} and Reg also learn a causal graph so that they can learn tasks with the state abstraction described in Sec. \ref{subsec:problem_definition}. Fig. \ref{fig:learning_curve} shows the training curves of the Lift and Stack tasks (those of other tasks are in appendix Sec. \ref{app:task_results}); \textsc{cdl} learns tasks the most sample-efficiently with the state abstraction. Though Reg also uses abstraction, it learns more slowly as some action-irrelevant state variables are also included in the abstraction because of the inaccurately learned causal graphs.

\begin{figure}
  \centering
  \includegraphics[width=1.0\columnwidth]{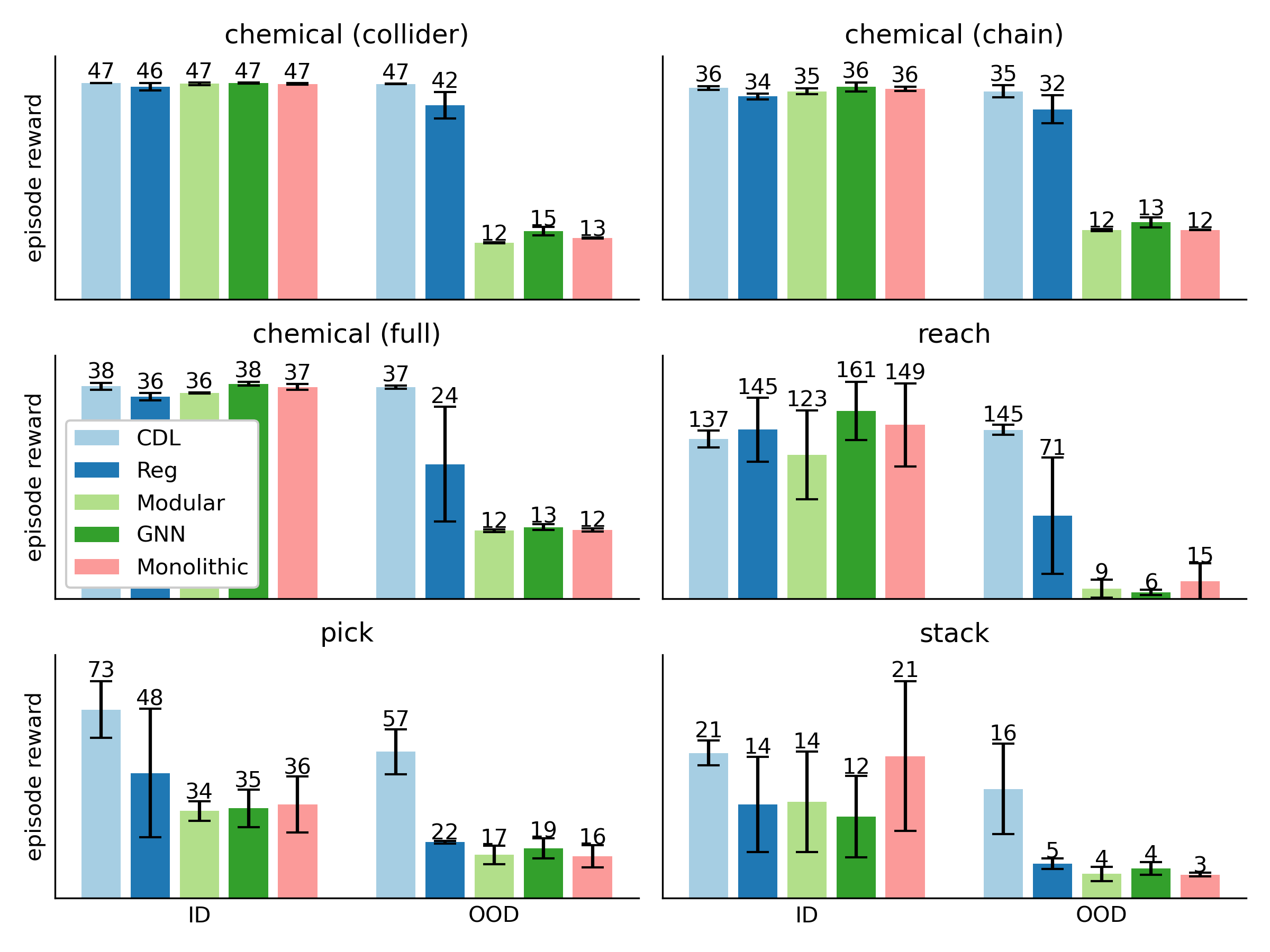}
  \vspace{-25pt}
  \caption{\small Episode reward of learned policies on ID and OOD states.}
  \label{fig:task_generalization}
  \vspace{-15pt}
\end{figure}

Furthermore, to test the generalization of the learned policies, we evaluate their performance for 100 episodes on both ID and OOD states, as shown in Fig. \ref{fig:task_generalization}. Again, \textsc{cdl} generalizes the best across all methods on OOD states.

\section{Conclusions}
\label{sec:conclusions}

In this work, we introduce \textsc{cdl}, Causal Dynamics Learning for Task-Independent State Abstraction. In contrast to most existing \textsc{mbrl} methods that learn a dense dynamics model, \textsc{cdl} learns and, for each state variable, only keeps necessary dependencies on other variables and actions, thus generalizing better to unseen states than dense models. 
Furthermore, \textsc{cdl} derives a state abstraction from the learned causal relationships, which can be applied to a wider range of tasks than existing state abstraction methods that only retain state variables specific to one task. 
Our experiments demonstrate \textsc{cdl}'s generalization improvement both on the learned dynamics models and the policies for downstream tasks, as well as its sample efficiency resulting from its state abstraction. 
While this paper focuses on low-dimensional state spaces, an important future direction is to extend \textsc{cdl} to high-dimensional space, like images.

\textbf{Acknowledgement}

This work has taken place in the Learning Agents Research
Group (LARG) at the Artificial Intelligence Laboratory, The University
of Texas at Austin.  LARG research is supported in part by the
National Science Foundation (CPS-1739964, IIS-1724157, FAIN-2019844),
the Office of Naval Research (N00014-18-2243), Army Research Office
(W911NF-19-2-0333), DARPA, General Motors, Bosch, and
Good Systems, a research grand challenge at the University of Texas at
Austin.  The views and conclusions contained in this document are
those of the authors alone.  Peter Stone serves as the Executive
Director of Sony AI America and receives financial compensation for
this work.  The terms of this arrangement have been reviewed and
approved by the University of Texas at Austin in accordance with its
policy on objectivity in research.

\bibliography{references}
\bibliographystyle{icml2022}

\newpage
\appendix

\onecolumn
\textbf{\Large Appendix for the paper: Causal Dynamics Learning for Task-Independent State Abstraction}

\section{Causal Dynamics Learning}
\subsection{Definitions}
\label{app:definitions}

Here we provide fundamental definitions in the causal inference that we use for proving Theorem. \ref{thm:causal_sufficient_condition}. For a thorough study see \citet{pearl2009causality}. 

\begin{definition}
\label{def:d_separation}
(d-separation \cite{pearl2009causality}) In a directed acyclic graph (DAG) $\mathcal{G}$, a path between nodes $I_1$ and $I_m$ is blocked by a set $S$ (with neither $I_1$ nor $I_m$ in $S$) whenever there is a node $I_k, k = 2, \dots,m - 1$, such that one of the following two possibilities holds:\\
(i) $I_k \in S$ and $I_{k-1} \rightarrow I_k \rightarrow I_{k+1}$ or $I_{k-1} \leftarrow I_k \leftarrow I_{k+1}$ or $I_{k-1} \leftarrow I_k \rightarrow I_{k+1}$. \\
(ii) Neither $I_k$ nor any of its descendants is in $S$ and $I_{k-1} \rightarrow I_k \leftarrow I_{k+1}$.
\end{definition}
In a DAG $\mathcal{G}$, we say that two nodes $A$ and $B$ are d-separated by a third node $C$ if every path between nodes $A$ and $B$ is blocked by $C$, denoted as $A \ind_\mathcal{G} B | C$.

\begin{definition}
\label{def:causal_markov_condition}
(Causal Markov Condition \cite{spirtes2000causation}) Let $\mathcal{G}$ be a causal graph with vertex set $V$ and $p$ be a probability distribution over the vertices in $V$ generated by the causal structure represented by $\mathcal{G}$. $\mathcal{G}$ and $p$ satisfy the Causal Markov Condition if and only if for every node $v$ in $V$, $v$ is independent of $V \setminus  (\text{Descendants}(v) \cup \text{Parents}(v))$ given Parents($v$).
\end{definition}

Here we use the global version of the Markov condition, which reads: if $A \ind_\mathcal{G} B | C \Rightarrow A \ind B | C$ for all disjoint vertex sets $A, B, C$ (where $\ind_\mathcal{G}$ denotes d-separation, as defined above, and $\ind$ denote independence in the probability).

\begin{definition}
\label{def:causal_faithfulness}
(Causal Faithfulness). A distribution $p$ is faithful to a DAG $\mathcal{G}$ if no conditional independence relations other than the ones entailed by the Markov property are present.
\end{definition}

\subsection{Proof of Theorem. \ref{thm:causal_sufficient_condition}}
\label{app:theorem_proof}

We need to show that if $s_t^i \not\to s_{t+1}^j$, then the condition $s_t^i \notind s^j_{t+1} | \{a_t, s_t \setminus s^i_t\}$ or the assumption is violated.

Assume there is no direct edge from $s_t^i$ to $s_{t+1}^j$ but they are dependent, i.e., $s_t^i \not\to s_{t+1}^j$ and $s_t^i \notind s_{t+1}^j$, then there is a confounder $Q_{t'}$ with the confounding path $s_t^i \dashleftarrow Q_{t'} \dashrightarrow s_{t+1}^j$ where $t' < t$ as we assume there is no simultaneous edge. There are two cases for $Q_{t'}$:\\
(1) If $Q_{t'}$ is observed, i.e., $Q_{t'}$ is one of the state variable, then it violates the condition because we condition on $\{a_t, s_t \setminus s^i_t\}$ which block all possible paths for $Q_{t'} \dashrightarrow s_{t+1}^j$ and thus d-separate $s_t^i$ and $s_{t+1}^j$. \\
(2) If $Q_{t'}$ is unobserved, then it must be memoryless (i.e., $Q_{t} \not\to Q_{t + 1}$) to adhere to A2 (the state is fully observed and the dynamics is Markovian). In that case, there exists such a path $s_{t-1}^i \rightarrow s_t^i \dashleftarrow Q_{t'} \dashrightarrow s_{t+1}^j$ given A3 ($s_t^i \rightarrow s^i_{t+1}$ exists for all state variables $s^i$). This path suggests $s_{t-1}^i \notind s_{t+1}^j | s_t^i$ where $s_t^i$ serves as the collider connecting $s_{t-1}^i$ and $s_{t+1}^j$. If we further condition the dependence on all other state variables at $t$ to block all potential paths $s_{t-1}^i \rightarrow s^k_{t} \rightarrow s_{t+1}^j$, we have $s_{t-1}^i \notind s_{t+1}^j | s_t$. However this conditional dependence violates A2, i.e., Markovian property of the dynamics $s_{t-1} \ind s_{t+1} | s_t$.

In both cases either the condition or A2 is violated. Therefore we show that if $s_t^i \notind s^j_{t+1} | \{a_t, s_t \setminus s^i_t\}$, then $s_t^i \rightarrow s^j_{t+1}$. \hfill$\square$

The proof of that if $a_t \notind s^j_{t+1} | s_t$, then $a_t \rightarrow s^j_{t+1}$ is even simpler: as $a_t$ has no parent in the causal graph, there cannot exist the confounding path $a_t \dashleftarrow Q_{t'} \dashrightarrow s_{t+1}^j$. Then following A1 (causal faithfulness), conditioning on all other potential parents of $s^j_{t+1}$ (i.e., $s_t$), the conditional dependence $a_t \notind s^j_{t+1} | s_t$ must result from the existence of the edge $a_t \rightarrow s^j_{t+1}$. \hfill$\square$

\subsection{Conditional Mutual Information}
\label{app:cmi_eq_derivation}

We provide the full derivation of Eq. \ref{eq:cmi} as follows:
\begin{align}
& \mathop{\mathbb{E}}\limits_{s_t,a_t,s^j_{t+1}}\left[\log\frac{p(s^i_t, s^j_{t+1} | \{a_t, s_t \setminus s^i_t\})}{p(s^i_t | \{a_t, s_t \setminus s^i_t\}) \ p(s^j_{t+1} | \{a_t, s_t \setminus s^i_t\})}\right] \nonumber\\
=& \mathop{\mathbb{E}}\limits_{s_t,a_t,s^j_{t+1}}\left[\log\frac{p(s^i_t | \{a_t, s_t \setminus s^i_t\}) \ p(s^j_{t+1} | \{a_t, s_t\})}{p(s^i_t | \{a_t, s_t \setminus s^i_t\}) \ p(s^j_{t+1} | \{a_t, s_t \setminus s^i_t\})}\right] 
\quad \textcolor{blue}{\text{// by expanding $p(s^i_t, s^j_{t+1} | \{a_t, s_t \setminus s^i_t\})$}}\nonumber\\
=& \mathop{\mathbb{E}}\limits_{s_t,a_t,s^j_{t+1}}\left[\log\frac{p(s^j_{t+1} | a_t, s_t)}{p(s^j_{t+1} | \{a_t, s_t \setminus s^i_t\})}\right].
\qquad \qquad \qquad \qquad \ \textcolor{blue}{\text{// by cancelling out $p(s^i_t | \{a_t, s_t \setminus s^i_t\})$}}\nonumber
\end{align}

\section{Additional Related Work}
\label{app:additional_related}

In this section, we given a more detailed comparison between our work and methods that consider explicit sparsity and modularity, e.g., Neural Production Systems (NPS, \citet{alias2021neural}) and Object Files and Schemata (SCOFF, \citet{goyal2020object}). Specifically:

1. \textbf{Sparsity}: Compared to our method in which each variable depends on all of its causal parents, NPS constrains each variable (i.e., ``slot" in NPS) to only depend on one parent. However, in many problems, the dynamics of one variable does depend on multiple variables, e.g., a hammer handed from one robot to the other robot, or the chemical environments in the paper with the collider graph. In those cases, NPS does not perform well due to its one-parent constraint, as shown in Table. 3. SCOFF assumes dense dependencies, so we leave it out of the sparsity discussion.

2. \textbf{Modularity}: NPS and SCOFF consider modular dynamics, i.e., multiple variables can share the same dynamics function, while our dynamics model does not permit such modularization. Extending the model accordingly is a good direction for future work.

3. \textbf{Global vs local dependencies}: Our method considers global dependencies that are static across all state values. In contrast, NPS's and SCOFF's local dependencies depend on the value of the state. \\
\underline{Cons of local dependencies}: Since the selection of causal parent depends on the value of state variables, out-of-distribution (OOD) variables may be assigned as the wrong parent, resulting in inaccurate prediction, as shown by OOD state evaluation in Table. 3. In contrast, in our method, dependencies are independent of variable values, and thus the prediction of variables will not be affected by OOD variables that are not their parents. \\
\underline{Pros of local dependencies}: They may induce an even sparser dynamics. For example, globally, an object depends on the robot hand which can move it, but, locally, it only depends on the hand when being grasped by the hand and is independent otherwise. As such, another promising future direction is to incorporate locality into global dependencies for further sparsity.

\begin{table}
\centering
\begin{small}
\caption{prediction accuracy (\%) in the chemical collider environment}
\begin{tabular}{ccc}
\toprule
\textbf{Method}         & \textbf{CDL} (Ours)   & \textbf{NPS} \\
\midrule
\textbf{in-distribution states} & \textbf{97.3}         & 88.0  \\
\textbf{OOD states}      & \textbf{97.3}         & 35.7  \\
\bottomrule
\end{tabular}
\end{small}
\vspace{-16pt}
\label{tab:nps_results}
\end{table}

\section{Experiment Details}
\subsection{Environment Details}
\label{app:environment}

\subsubsection{Chemical Environment}
In the chemical environment modified from \cite{ke2021systematic}, there are 10 objects and each of them has its different x, y position (two continuous variables) and color (a categorical variable out of 5 possible colors), forming 30 state variables in total.
At each step, the x, y positions of objects are sampled from a normal distribution $\mathcal{N}(0, 1e^{-4})$. Meanwhile, the action changes a chosen object to a specified new color (a categorical variable out of 50 possible actions). After the chosen object changes its color, all its decadents in the underlying graph will change their colors in the breadth-first order. When changing to its new color, each descendant follows a pre-defined conditional probability table modeled as a randomly initialized multi-layer perceptron (MLP). The MLP takes the newest color of each parent as inputs and outputs a categorical distribution over 5 colors out of which one color is sampled for the object. For example, in the chain graph shown in Fig. \ref{fig:chemical_env} (a) top, if the color of the leftmost object is changed by the action, then the second left object will change its color given the new color of the leftmost object changes to, then the third object changes, etc.

In the ground truth causal graph, each object's x position or y position only depends on itself and thus is action-irrelevant, while the color depends on the action and the colors of its parents and thus is controllable. However, if an object only has one parent, then its color doesn't depend on the color of that parent, as whenever its color needs to change, it's either (1) because the color is directly set by the action where the parent color has no effect or; (2) because that its only parent changes the color, and this new parent color it is going to depend on is not included in $s_t$ ($s_t$ only includes the parent color before the change). As a result, there is no dependence from the parent color at $t$ to its new color at $t + 1$.

\subsubsection{Manipulation Environment}

In the manipulation environment, the x-axis is along the width of the table, the y-axis is along the length, and z-axis is along the height. If we set the center of the table top as $(x, y, z)=(0, 0, 0)$, at each episode, the positions of all objects are uniformly sampled from $(x, y) \sim \text{Uniform}([-0.3, 0.3] \times [-0.4, 0.4])$ without overlapping and markers are sampled from $(x, y) \sim \text{Uniform}([-0.35, 0.35] \times [-0.5, 0.5])$. The end-effector of the robot is constrained in a box with the range of $[-0.3, 0.3] \times [-0.4, 0.4]\times [0, 0.2]$. At each step, the position change of the randomly moving object is sampled from $\mathcal{N}(0, 4e-6)$ and the changes of markers are sampled from $\mathcal{N}(0, 1e-4)$, and they are constrained in their initialization ranges.

\begin{figure}
  \centering
  \includegraphics[width=1.0\columnwidth]{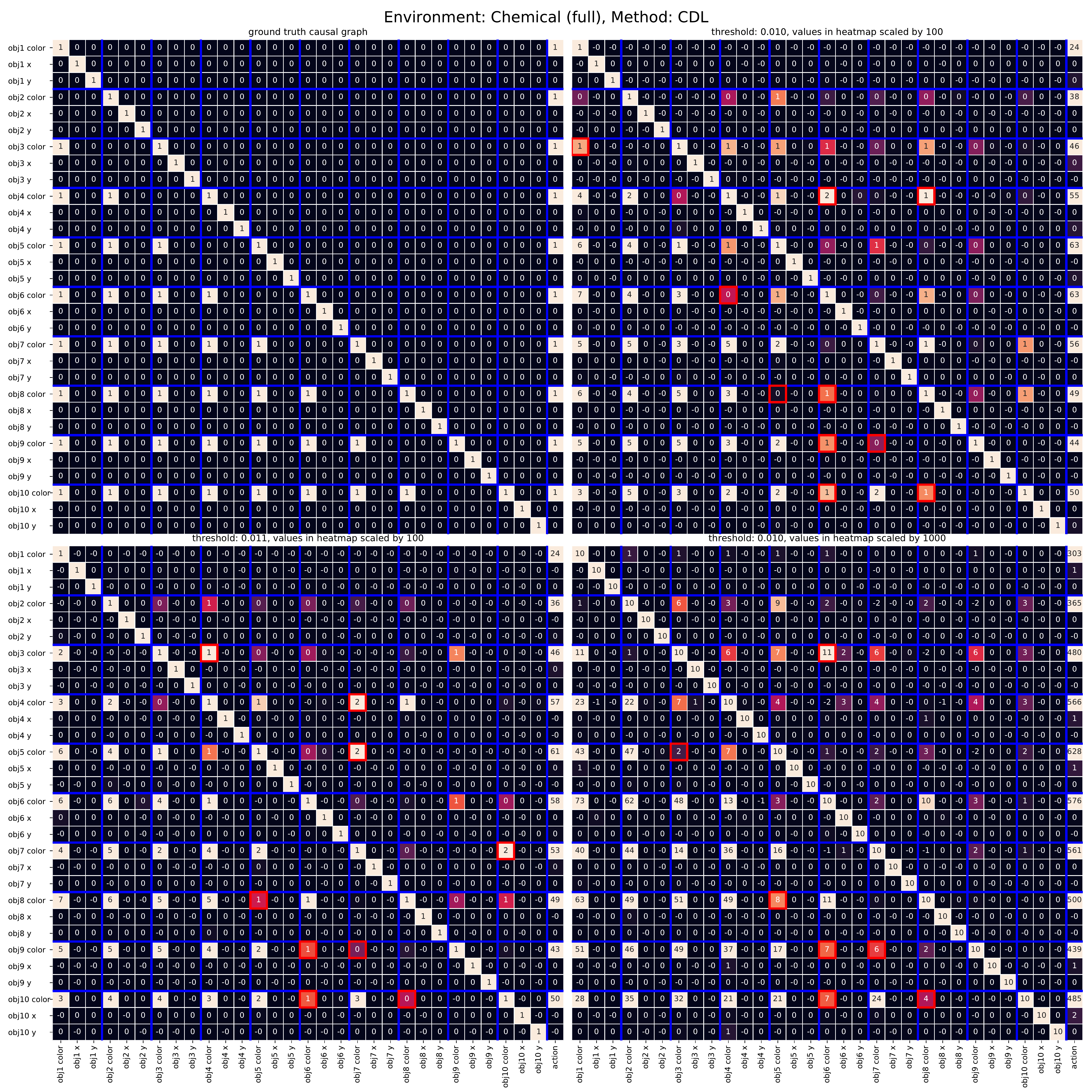}
  \vspace{-20pt}
  \caption{\small Causal graph for the chemical environment with the full graph. \textbf{(top left)} the ground truth causal graph, \textbf{(others)} causal graphs learned by \textbf{\textsc{cdl}} with 3 different runs. Among the three graphs for the chemical environment, the dense graph is the most complex one, with both long-term dependencies and much more dependencies to capture. As a result, \textsc{cdl} makes more prediction errors, but the learn the causal graphs are still much more accurate than those learned by Reg.}
  \label{fig:causal_graph_full_cmi}
  \vspace{-15pt}
\end{figure}

In the ground truth causal graph shown in Fig. \ref{fig:causal_graph_full_cmi} top left, the controllable state variables include EEF position, gripper joint angles, and movable object positions; the action-relevant state variables are positions of the unmovable object and the randomly moving object as they can block the motion of the EEF, gripper, and the movable object (when pushed or held by the gripper); and action-irrelevant state variables include positions of all non-interactable markers.

\subsubsection{Physical Environment}

In addition to the chemical and manipulation environment, we also evaluate our method in the physical environment proposed by \cite{ke2021systematic}. In a $5 \times 5$ grid-world, there are 5 objects and each of them has a unique weight. The state space is 10-dimensional, consisting of x, y positions (a categorical variable over 5 possible values) of all objects. At each step, the action selects one object, moves it in one of 4 directions or lets it stay at the same position (a categorical variable over 25 possible actions). During the movement, only the heavier object can push the lighter object (the object won't move if it tries to push an object heavier than itself). Meanwhile, the object cannot move out of the grid-world nor can it push other lighter objects out of the grid-world. Moreover, the object cannot push two objects together, even when both of them are lighter than itself.

For the ground truth causal graph, though the original authors state that there are only edges from heavier objects to lighter objects, we want to point out that the graph is actually a dense graph that also includes edges from lighter objects to heavier objects, because lighter objects can also block the movement of heavier objects if they are at the boundary or two lighter ones block one heavier object together, as described above. However, in the physical environment, as all state variables are controllable and each state variable depends on all other variables, causal dynamics learning would learn a dense model and thus has no advantages over dense dynamic models.

\subsection{Implementation Details}

\subsubsection{Architectures}
\label{app:architectures}
We listed the dynamics model architectures of our method (\textsc{cdl}) and all baselines in Table. \ref{tab:architecture} (note that \textsc{cdl}, Reg, and Modular share the same architecture) and hyperparameters shared across methods in Table. \ref{tab:general_hyperparams}. Meanwhile, the same architecture and hyperparameters are used for the chemical environment with different underlying graphs. For the MLP network, a list is used to represent its architecture, e.g., [64, 128] represents an MLP of 2 hidden layers with 64 neurons in the first layer and 128 neurons in the second. Notice that the input layer and the output layer of the MLP are not listed, so an empty list [] represents a linear mapping from inputs to outputs. For all activation functions, ReLU is used.

\begin{table}
\centering
\caption{Architectures for \textsc{cdl} and Baselines in All Environments}
\vspace{-0pt}
\begin{small}
\begin{tabular}{clccc}
\toprule
\multicolumn{2}{c}{\textbf{Architectures}}              & \multicolumn{3}{c}{\textbf{Environments}} \\
\midrule
Methods     & \multicolumn{1}{c}{Modules}               & Chemical  & Manipulation  & Physical\\
\midrule
\multirow{3}{*}{\textbf{\textsc{cdl}}, \textbf{Reg}, \textbf{Modular}} 
& feature extractors, $\{f^a_j, f^{1:\dS}_j\}_{j=1}^\dS$    & []            & []            & []            \\
& feature dimension, $h_{1:\dS}$                            & 64            & 128           & 128           \\
& predictive networks, $q_{1:\dS}$                          & [64, 32]      & [128, 64]     & [128, 128]    \\
\midrule
\multirow{6}{*}{\textbf{GNN}} 
& embedders (map state variables to node attributes)        & []            & []            & []            \\
& node attribute dimension                                  & 256           & 256           & 256           \\
& edge attribute dimension                                  & 256           & 256           & 256           \\
& node networks                                             & [256, 256]    & [256, 256]    & [256, 256]    \\
& edge networks                                             & [256, 256]    & [256, 256]    & [256, 256]    \\
& projects (map node attributes to state variables)         & []            & []            & []            \\
\midrule
\textbf{Monolithic} & MLP network                           & [256, 256, 256]   & [512, 512, 512]   & [256, 256, 256]   \\
\bottomrule
\end{tabular}
\end{small}
\vspace{-0pt}
\label{tab:architecture}
\end{table}

\begin{table}
\centering
\caption{Hyperparameters Shared across All Methods and Environments if not Specified}
\vspace{-0pt}
\begin{small}
\begin{tabular}{cccc}
\toprule
\multirow{2}{*}{\textbf{Name}}          & \multicolumn{3}{c}{\textbf{Environments}} \\
                                        & Chemical  & Manipulation  & Physical      \\
\midrule
number of transitions                   & 100K      & 28.8M         & 2M            \\
training step                           & 500K      & 1.8M          & 2M            \\
optimizer                               & \multicolumn{3}{c}{Adam}                  \\
learning rate                           & \multicolumn{3}{c}{3e-4}                  \\
batch size                              & \multicolumn{3}{c}{32}                    \\
prediction step during training, $H$    & \multicolumn{3}{c}{3}                     \\
\bottomrule
\end{tabular}
\end{small}
\vspace{-0pt}
\label{tab:general_hyperparams}
\end{table}

\subsubsection{Dynamics Learning Implementation Details}
\label{app:dynamics_learning}

For all methods, during training, instead of optimizing 1-step prediction loss, we roll out the model for $H$ steps and optimize the sum of $H$-step prediction loss. Also, we keep $10\%$ of the data as the validation data to select the best model for each method. Moreover, \textsc{cdl} also evaluate the conditional mutual information (\textsc{cmi}) using the validation data rather than training data for more accurate measurements. For Reg whose performance is sensitive to the regularization coefficient, we conduct a grid search for the best coefficients in terms of the causal graph accuracy, and its value for each environment is listed in Table. \ref{tab:cdl_dynamics_hyperparams}.

During \textsc{cdl} training, for the continuous state variable where \textsc{cdl} predicts its next value as the mean and standard deviation of a normal distribution, we find that clipping the range of standard deviation between $[0.001, 0.01]$ reduces overfitting and apply it to other baselines as well. For \textsc{cmi} evaluation, as it's expensive to evaluate it using all validation data whenever the dynamics model is updated, we instead sample a batch of transitions from the validation data and evaluate \textsc{cmi} using it once every 10 updates of dynamics parameters $\theta$. To get a smooth estimate of \textsc{cmi} while assigning greater weights to estimates from the newest model, we apply an exponential moving average to the \textsc{cmi} calculated from each batch. All hyperparameters of causal dynamics learning are listed in Table. \ref{tab:cdl_dynamics_hyperparams}.

\begin{table}
\centering
\caption{Hyperparameters for Dynamics Learning of \textsc{cdl} and Reg (Shared Across Environments if not Specified)}
\vspace{-0pt}
\begin{small}
\begin{tabular}{ccccccc}
\toprule
\textbf{Method} & \textbf{Name}                 & \multicolumn{5}{c}{\textbf{Environments}} \\
                &                               & \thead{Chemical\\(collider)}  
                                                & \thead{Chemical\\(chain)}  
                                                & \thead{Chemical\\(full)}  
                                                & Manipulation  & Physical    \\
\midrule
\multirow{5}{*}{\textsc{cdl}}
& \textsc{cmi} threshold, $\epsilon$            & 0.02      & 0.05      & 0.01      & 0.002     & 0.01     \\
& $\hat{p}(s^j_{t+1} | \textbf{PA}_{s^j})$ optimization frequency     & \multicolumn{5}{c}{once every 10 updates of $\theta$}     \\
& \textsc{cmi} evaluation frequency                                   & \multicolumn{5}{c}{once every 10 updates of $\theta$}     \\
& \textsc{cmi} evaluation batch size                                  & \multicolumn{5}{c}{32}                                    \\
& exponential moving average discount                                 & \multicolumn{5}{c}{0.999}                                 \\
\midrule
\multirow{2}{*}{Reg}
& regularization coefficient                    & 0.002     & 0.002     & 0.002     & 0.001     & 0.0     \\
& regularization starts after N training steps  & 100K      & 100K      & 100K      & 750K      & 0     \\
\bottomrule
\end{tabular}
\end{small}
\vspace{-0pt}
\label{tab:cdl_dynamics_hyperparams}
\end{table}

\subsubsection{Transition Collection Policy Implementation Details}
\label{app:transition_collection_policy_implementation}

For the chemical environment where the state-action space is simple, we just use a random policy that uniformly samples actions over 50 possible values. For the physical environment where the interactions between objects are infrequent, we use a scripted policy that iteratively selects the object to move and the object to push with the former object so that it covers all iterations between objects. For the chemical and physical environment, all methods are trained with the same transitions collected by policies mentioned earlier.

For the manipulation environment where efficient exploring is still a challenging research topic and not the focus of this paper, to reduce the complexity of exploration, we follow \citet{nasiriany2021maple} and modify the action space of the RL agent to following long-horizon skills: \\
(1) Reach: move the end-effector to the position of a chosen object or a target position$(x, y, z)$ . Execution takes at most 50 atomic actions. \\
(2) Grasp: move the end-effector to the position of a chosen object or a target position $(x, y, z)$. Execution takes at most 50 atomic actions. \\
(3) Lift: lift one of the objects to a target position $(x, y, z)$. Execution takes at most 50 atomic actions. \\
(4) Push: select one of the objects whose position is $(x, y, z)$ and then move the end-effector from $(x + \delta_x, y + \delta_y, z + \delta_z)$ to $(x + \Delta_x, y + \Delta_y, z + \Delta_z)$. Execution takes at most 50 atomic actions.\\
(5) Open: open the gripper. Execution takes at most 4 atomic actions. \\
(6) Atomic: apply a single atomic action, which allows the agent to fill in any gaps that cannot be fulfilled by other skills.

Notice that dynamics learning still uses low-level raw motor actions (i.e., atomic actions). To train the agent whose action space becomes low-frequency high-level skills but whose rewards are still computed from high-frequency low-level atomic actions, we adopt HiPPO \cite{li2019sub}. Similar to PPO, in HiPPO, the critic uses the current state and atomic action as inputs and is trained to fit state values computed from rewards, and the actor is trained to maximize the advantage but only at timestamps when it selects skills. The actor consists of (1) a skill selection network that selects one of six skills based on $s_t$, (2) an object selection network that selects the object to apply the skill to based on $s_t$ and the selected skill, and (3) a skill parameter network that outputs skill parameter (e.g., goal position $(x, y, z)$) based on $s_t$, the selected skill and the selected object. If the selected skill does not apply to an object or does not need parameters, the selected object or skill parameters will be ignored. The architecture and hyperparameters of HiPPO are listed in Table. \ref{tab:hippo_hyperprarams}. For the manipulation environment, for a fair comparison, all methods are trained with the same transitions that the HiPPO agent collects for \textsc{cdl} training.

\begin{table}
\centering
\caption{Architecture and Hyperparameters for Transition Collection Policy in the Manipulation Environment}
\vspace{-0pt}
\begin{small}
\begin{tabular}{cc}
\toprule
\textbf{Name}   & \textbf{Value} \\
\midrule
critic network              & [128, 128] \\
skill selection network     & [64, 64] \\
object selection network    & [64, 64] \\
skill parameter network     & [64, 64] \\
\midrule
optimizer                   & Adam \\
learning rate               & 1e-4 \\
batch size                  & 32 \\
discount                    & 0.95 \\
generalized advantage estimator parameter, $\lambda$    & 0.98 \\
clip ratio                  & 0.1 \\
entropy weight              & 0.2 \\
\bottomrule
\end{tabular}
\end{small}
\vspace{-0pt}
\label{tab:hippo_hyperprarams}
\end{table}

\subsection{Causal Dynamics Learning Evaluation Details}
\label{app:dynamics_results}

Apart from the results shown in Sec. \ref{subsec:dynamics_results}, we also evaluate the dynamics model learned by $\textsc{cdl}$ and baselines with other metrics and experiments which we show their results in this section.

\subsubsection{Causal Relationship Inference Details}
\label{app:causal_graph_results}

For $\textsc{cdl}$, after training the dynamics model and evaluating \textsc{cmi}, we derive the causal graph by examining whether $\textsc{cmi} \ge \epsilon$ for each edge. The threshold $\epsilon$ is an important hyperparameter that determines the accuracy of the derived causal graph and thus the accuracy of the dynamics model. Even though we can use the same value used for training, as the best threshold is unknown before training, that value is just roughly estimated from several runs and may not be the best threshold. As a result, to select the best threshold, we conduct a linear search and select the one that gives the best F1 score. For a fair comparison, we apply the same threshold selection method to the regularization-based causal dynamics learning (Reg): Reg learns each edge of the causal graph as an individual Bernoulli distribution where its success probability is the learnable parameter. During training, as a rule-of-thumb, we use 0.5 as the threshold when judging whether an edge exists for each sample. After training and when evaluating the causal graph based on the learned success probability, we conduct a linear search and again select the threshold that gives the best F1 score.

\begin{table}
\centering
\caption{Performance on Causal Graph Learning for \textsc{cdl} and Reg on All Environments}
\vspace{-0pt}
\begin{small}
\begin{tabular}{ccccccc}
\toprule
\textbf{Metrics} & \textbf{Method}  & \multicolumn{5}{c}{\textbf{Environments}} \\
                 &                  & \thead{Chemical\\(collider)}  
                                    & \thead{Chemical\\(chain)}  
                                    & \thead{Chemical\\(full)}  
                                    & Manipulation  & Physical    \\
\midrule
\multirow{2}{*}{\textbf{Accuracy}}
& \textsc{cdl}                      & \textbf{1.000} $\pm$ 0.000   & \textbf{1.000} $\pm$ 0.001   & \textbf{0.991} $\pm$ 0.001   & \textbf{0.902} $\pm$ 0.003   & 1.00 $\pm$ 0.000   \\
& Reg                               & 0.994 $\pm$ 0.004            & 0.997 $\pm$ 0.001            & 0.977 $\pm$ 0.004            & 0.844 $\pm$ 0.005            & 1.00 $\pm$ 0.000   \\
\midrule
\multirow{2}{*}{\textbf{Recall}}
& \textsc{cdl}                      & \textbf{1.000} $\pm$ 0.000   & \textbf{0.992} $\pm$ 0.012   & \textbf{0.917} $\pm$ 0.016   & \textbf{0.612} $\pm$ 0.009   & 1.00 $\pm$ 0.000   \\
& Reg                               & 0.891 $\pm$ 0.086            & 0.942 $\pm$ 0.012            & 0.794 $\pm$ 0.006            & 0.459 $\pm$ 0.048            & 1.00 $\pm$ 0.000   \\
\midrule
\multirow{2}{*}{\textbf{Precision}}
& \textsc{cdl}                      & 1.000 $\pm$ 0.000            & 1.000 $\pm$ 0.000            & \textbf{0.968} $\pm$ 0.016   & \textbf{0.980} $\pm$ 0.006   & 1.00 $\pm$ 0.000   \\
& Reg                               & 0.993 $\pm$ 0.010            & 0.991 $\pm$ 0.012            & 0.917 $\pm$ 0.051            & 0.839 $\pm$ 0.068            & 1.00 $\pm$ 0.000   \\
\midrule
\multirow{2}{*}{\textbf{F1 Score}}
& \textsc{cdl}                      & \textbf{1.000} $\pm$ 0.000   & \textbf{0.996} $\pm$ 0.006   & \textbf{0.941} $\pm$ 0.009   & \textbf{0.754} $\pm$ 0.008   & 1.00 $\pm$ 0.000   \\
& Reg                               & 0.937 $\pm$ 0.046            & 0.966 $\pm$ 0.006            & 0.850 $\pm$ 0.019            & 0.589 $\pm$ 0.025            & 1.00 $\pm$ 0.000   \\
\midrule
\multirow{2}{*}{\textbf{ROC AUC}}
& \textsc{cdl}                      & \textbf{1.000} $\pm$ 0.000   & 0.996 $\pm$ 0.006            & 0.958 $\pm$ 0.008            & 0.805 $\pm$ 0.005            & 1.00 $\pm$ 0.000   \\
& Reg                               & 0.932 $\pm$ 0.042            & 0.990 $\pm$ 0.009            & 0.956 $\pm$ 0.008            & \textbf{0.893} $\pm$ 0.002   & 1.00 $\pm$ 0.000   \\
\bottomrule
\end{tabular}
\end{small}
\vspace{-0pt}
\label{tab:causal_graph_quality}
\end{table}

After deriving the causal graph for each run with its best threshold, we evaluate the performance of \textsc{cdl} and Reg on causal graph learning, in terms of accuracy, recall, precision, F1 score, and area under the receiver operating characteristic curve (ROC AUC). For all metrics, the higher the better. The results for different environments are shown in Table. \ref{tab:causal_graph_quality}. Again, $\textsc{cdl}$ outperforms Reg for most of the metrics and environments, except for ROC AUC in the manipulation environments. In the physical environment, both methods learn the dense causal graph correctly and thus have values of 1.000 for all metrics.

\begin{figure}
  \centering
  \includegraphics[width=1.0\columnwidth]{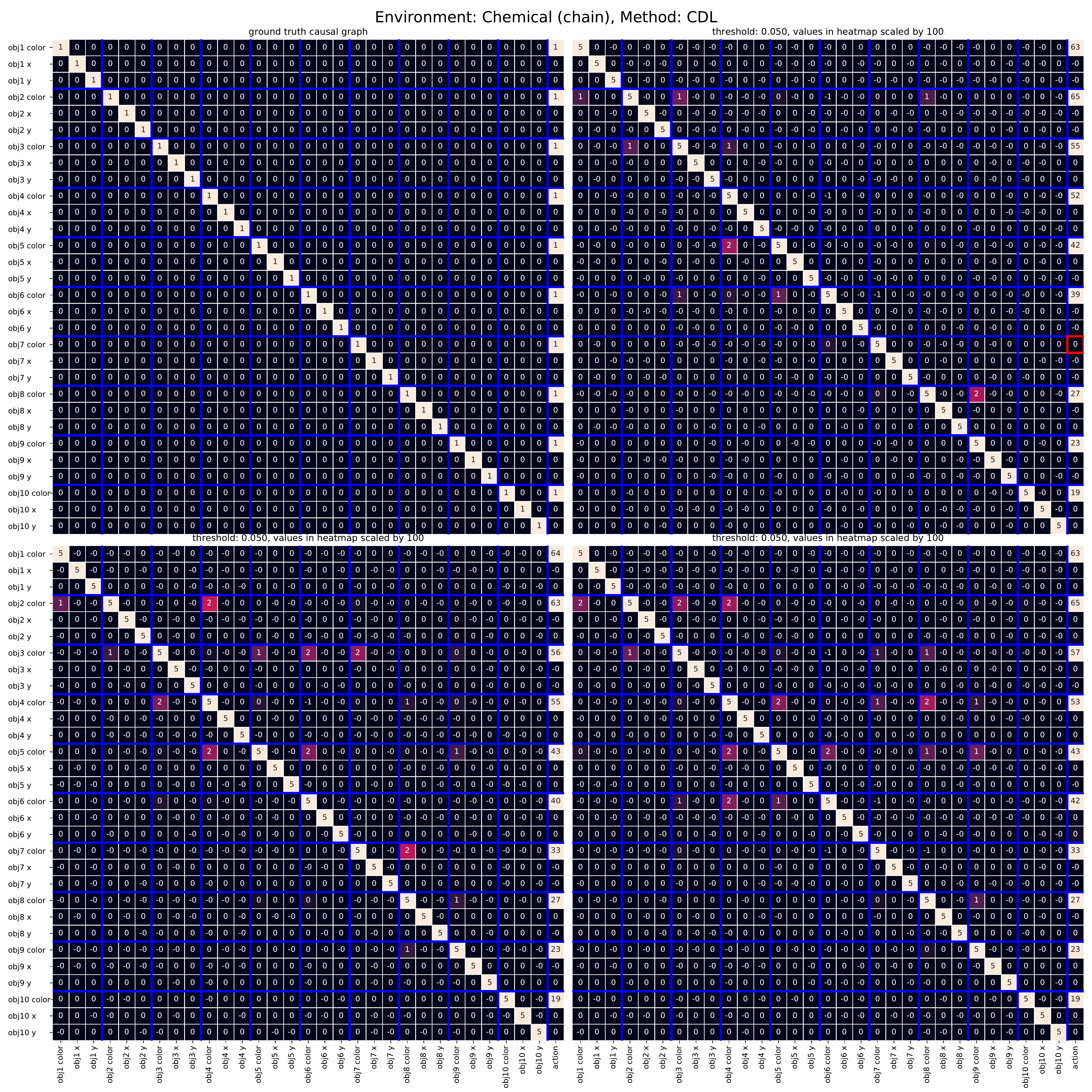}
  \vspace{-20pt}
  \caption{\small Causal graph for the chemical environment with the chain graph. \textbf{(top left)} the ground truth causal graph, \textbf{(others)} causal graphs learned by \textbf{\textsc{cdl}} with 3 different runs. Despite that the underlying graph is more challenging than the collider graph with long-term depdencies, all three runs still learn the causal graph accurately except for one missing dependency in top right.}
  \label{fig:causal_graph_chain_cmi}
  \vspace{-15pt}
\end{figure}

\begin{figure}
  \centering
  \includegraphics[width=1.0\columnwidth]{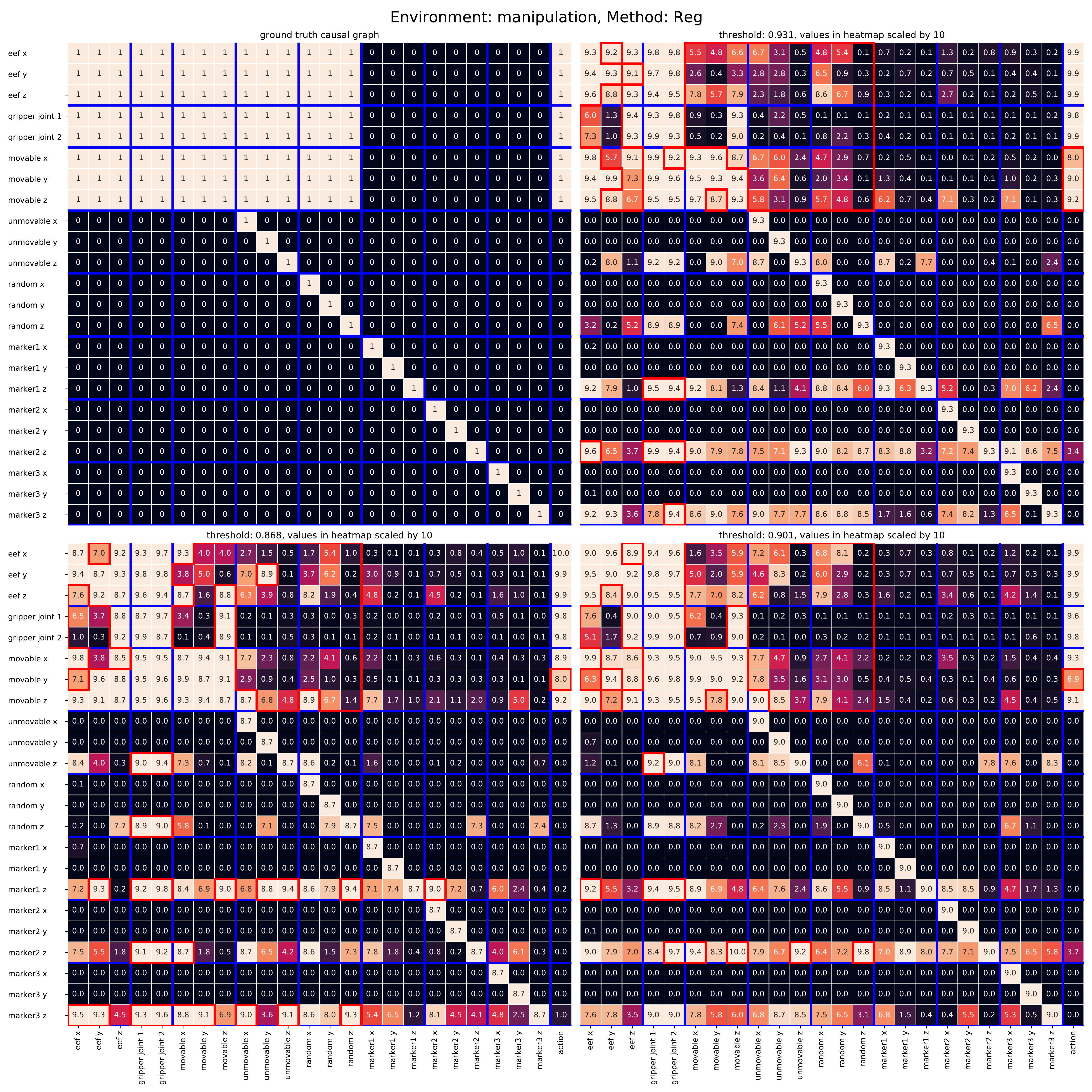}
  \vspace{-20pt}
  \caption{\small Causal graph for the manipulation environment. \textbf{(top left)} the ground truth causal graph, \textbf{(others)} causal graphs learned by \textbf{Reg} with 3 different runs, which clearly illustrate the challenging tradeoff between prediction and regularization for regularization-based method. Even though trained under the same regularization coefficient, frequent events, like the movement of markers, use lots of spurious correlations to improve prediction performance, while rare events, like eef's dependency on the unmovable objects which may block it, are sacrificed for lower regularization penalty. Consequently, the causal graphs learned by Reg have many false positives and false negatives at the same time.}
  \label{fig:causal_graph_manipulation_reg}
  \vspace{-15pt}
\end{figure}

In Fig. \ref{fig:causal_graph_chain_cmi} $\sim$ \ref{fig:causal_graph_manipulation_reg}, we also show the causal graph learned by both methods in all environments except for the physical environment as both methods learn it equally well. In each quadruplet figure, the top left one shows the ground truth causal graph of the environment which is binary, while the other three show the causal graph learned by one of the methods with three runs. In each causal graph, there are $\dS$ rows and $\dS + 1$ columns, and the element at the $j$-th row and $i$-th column represents whether the variable $s_{t+1}^j$ depends on the variable $s_t^i$ if $j < \dS + 1$ or $a_t$ if $j = \dS + 1$, measured by \textsc{cmi} for \textsc{cdl} or Bernoulli success probability for Reg. As \textsc{cmi} and success probability are continuous values, we plot them as a heatmap where darker color represents smaller values and vice versa. Also, the color in each heatmap is capped at the threshold selected as described in Sec. \ref{app:dynamics_results} so that all inferred causal edges are shown in the same white (brightest) color, and the selected threshold is marked on top of each heatmap. Finally, we mark all wrong edge predictions (i.e., false positives and false negatives) using red boxes for easier identification.

\begin{figure}
  \centering
  \includegraphics[width=1.0\columnwidth]{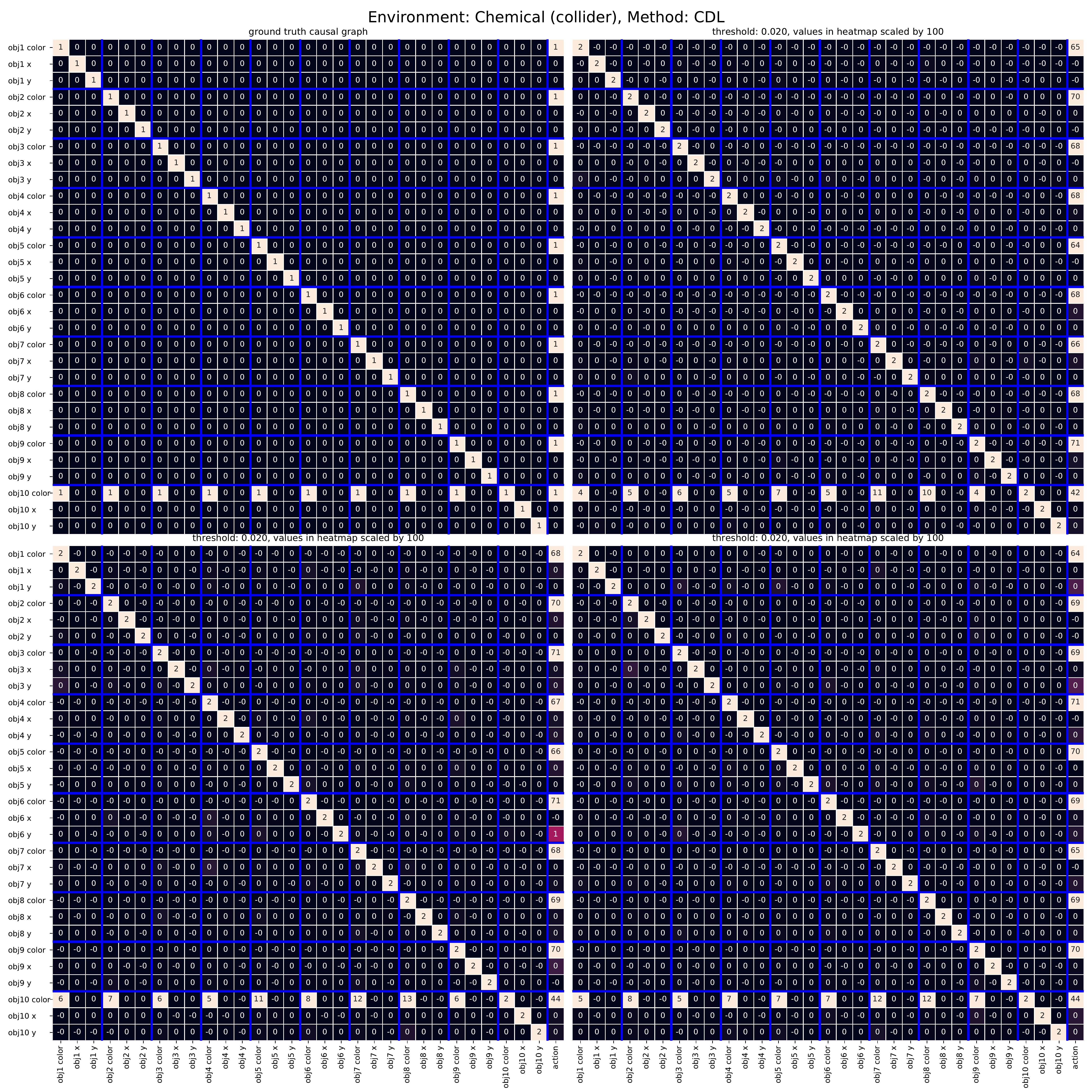}
  \vspace{-20pt}
  \caption{\small Causal graph for the chemical environment with the collider graph. \textbf{(top left)} the ground truth causal graph, \textbf{(others)} causal graphs learned by \textbf{\textsc{cdl}} with 3 different runs, where all of them learn the causal graph accurately.}
  \label{fig:causal_graph_collider_cmi}
  \vspace{-15pt}
\end{figure}

\begin{figure}
  \centering
  \includegraphics[width=1.0\columnwidth]{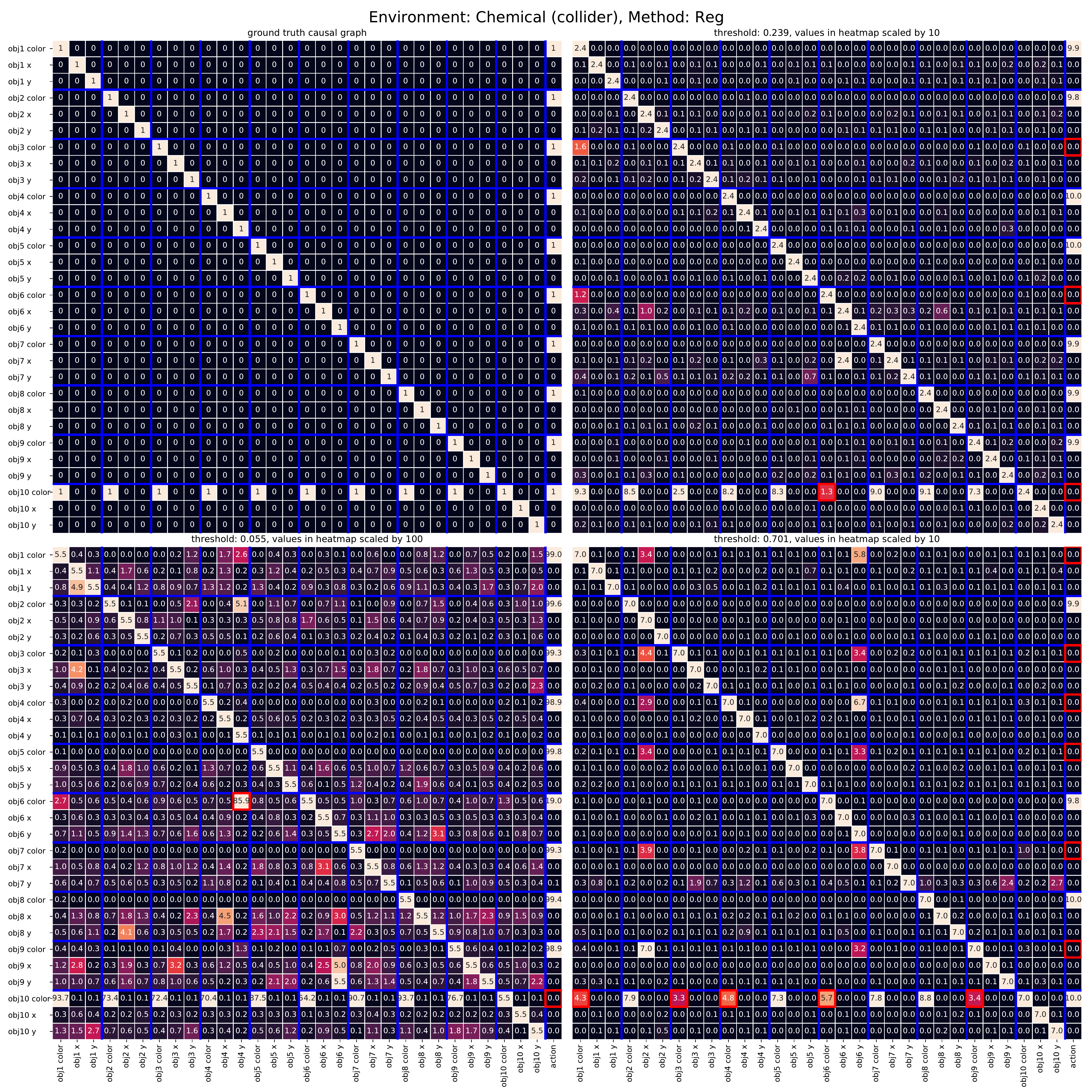}
  \vspace{-20pt}
  \caption{\small Causal graph for the chemical environment with the collider graph. \textbf{(top left)} the ground truth causal graph, \textbf{(others)} causal graphs learned by \textbf{Reg} with 3 different runs, where there are multiple missing dependencies (false negatives) and spurious correlations (false positives).}
  \label{fig:causal_graph_collider_reg}
  \vspace{-15pt}
\end{figure}

\begin{figure}
  \centering
  \includegraphics[width=1.0\columnwidth]{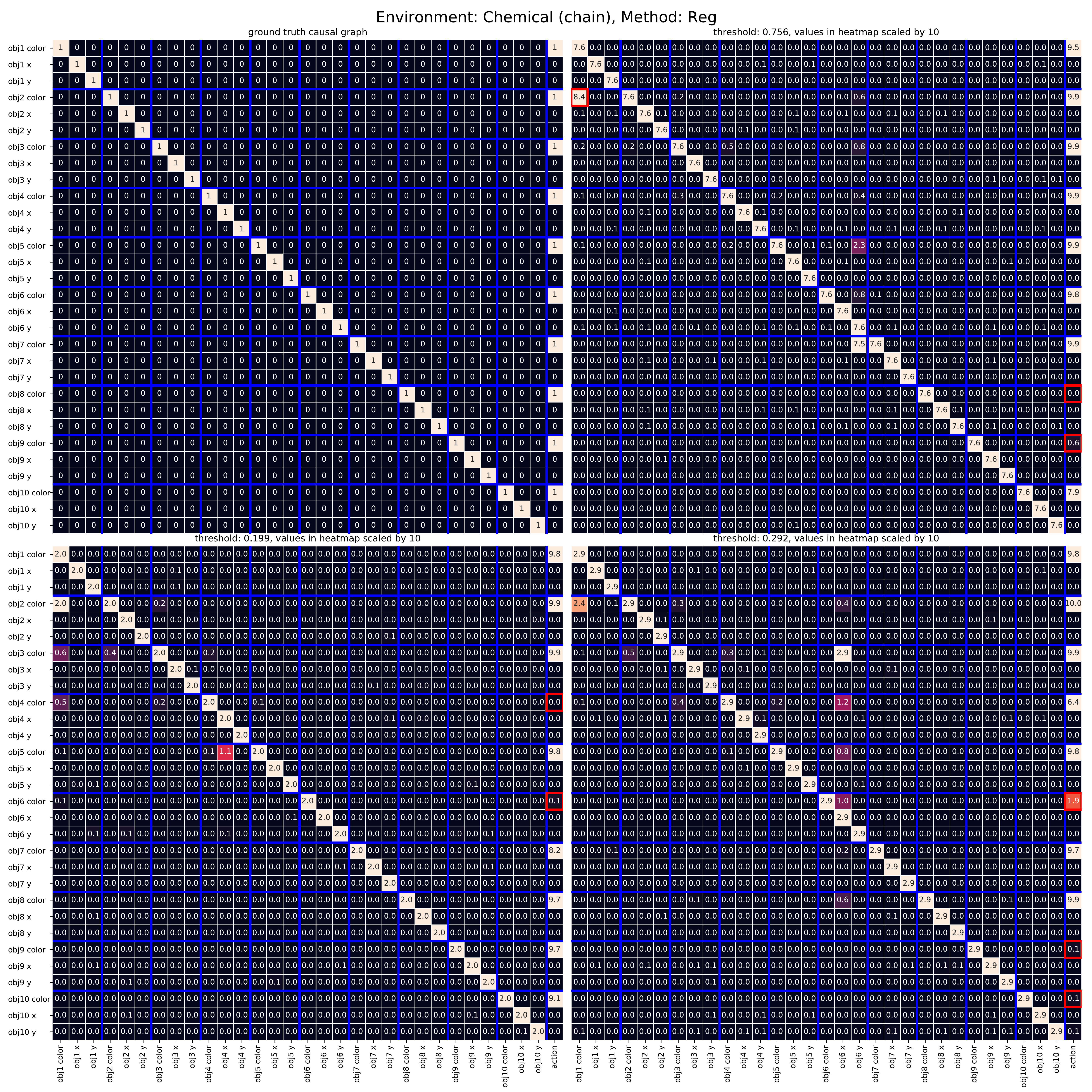}
  \vspace{-20pt}
  \caption{\small Causal graph for the chemical environment with the chain graph. \textbf{(top left)} the ground truth causal graph, \textbf{(others)} causal graphs learned by \textbf{Reg} with 3 different runs, where there again are multiple missing dependencies (false negatives) and spurious correlations (false positives) in all runs. Furthermore, Reg misses several dependencies on the action which are crucial to downstream task learning.}
  \label{fig:causal_graph_chain_reg}
  \vspace{-15pt}
\end{figure}

\begin{figure}
  \centering
  \includegraphics[width=1.0\columnwidth]{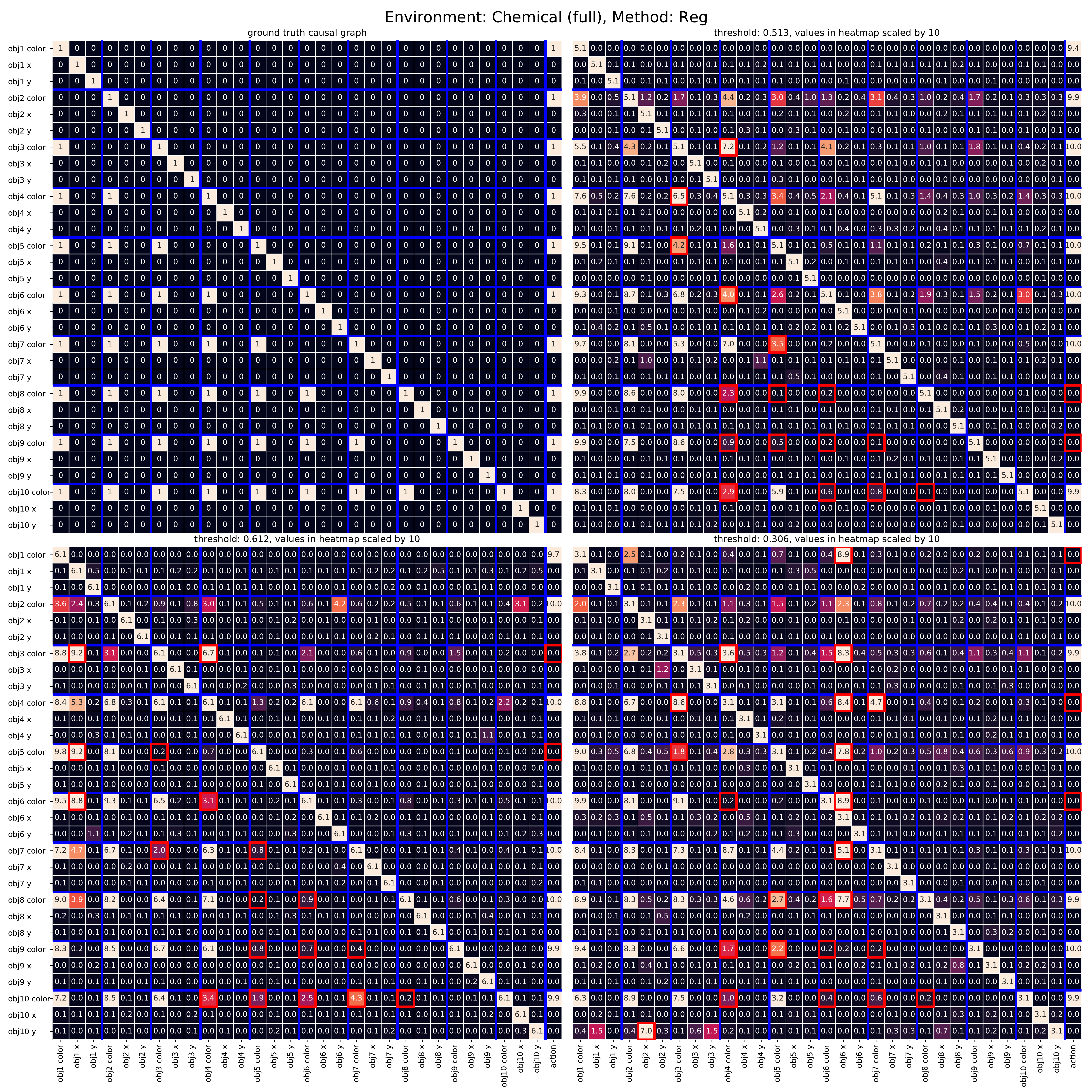}
  \vspace{-20pt}
  \caption{\small Causal graph for the chemical environment with the full graph. \textbf{(top left)} the ground truth causal graph, \textbf{(others)} causal graphs learned by \textbf{Reg} with 3 different runs, where there are much more missing dependencies (false negatives) and spurious correlations (false positives) than the graphs learned by \textsc{cdl}.}
  \label{fig:causal_graph_full_reg}
  \vspace{-15pt}
\end{figure}

\begin{figure}
  \centering
  \includegraphics[width=1.0\columnwidth]{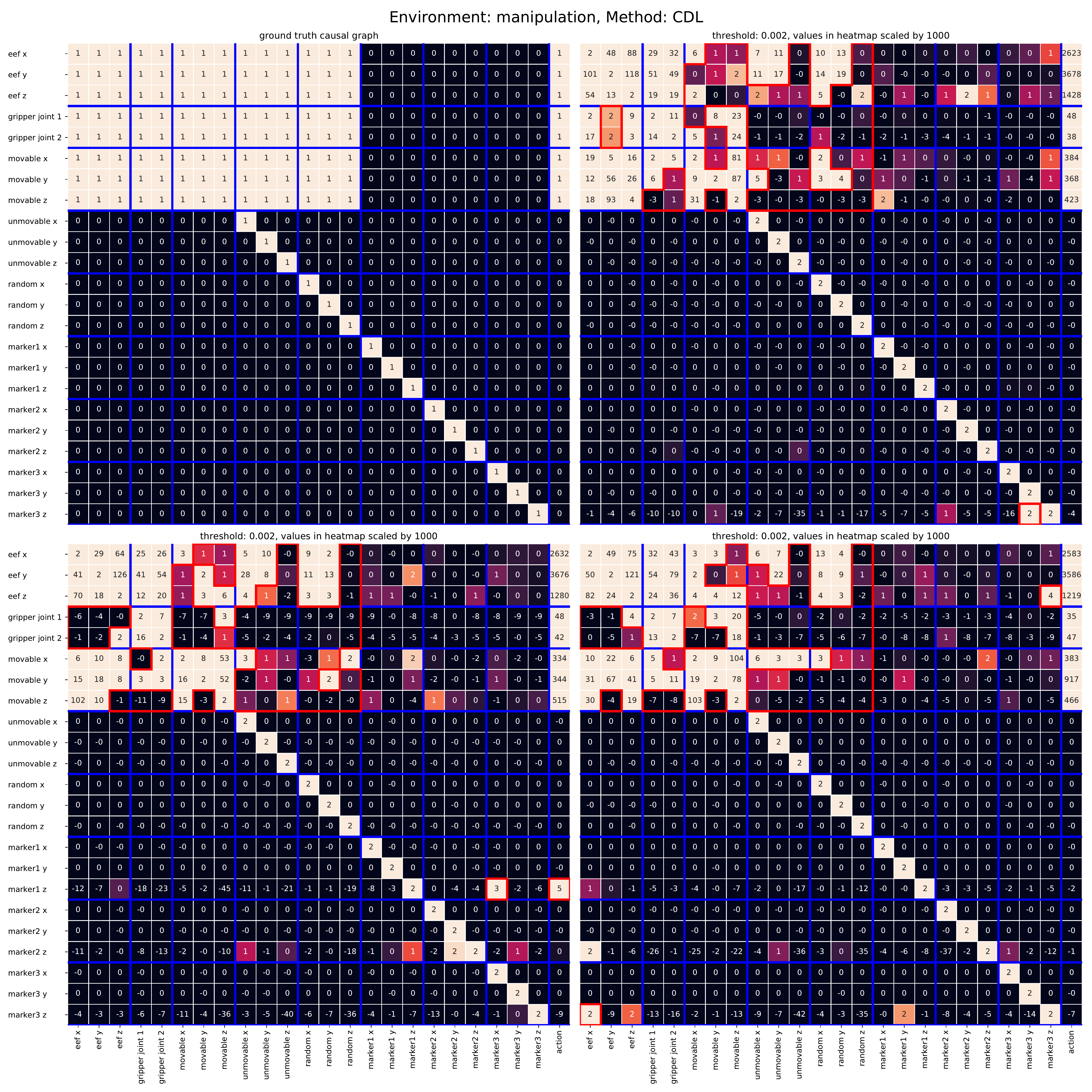}
  \vspace{-20pt}
  \caption{\small Causal graph for the manipulation environment. \textbf{(top left)} the ground truth causal graph, \textbf{(others)} causal graphs learned by \textbf{\textsc{cdl}} with 3 different runs. In all runs, though \textsc{cdl} successfully learns the controllable variables: eef, gripper, and movable object, it only captures part of their dependencies on the unmovable and the randomly moving object as the interactions between them are infrequent. Also, it gives one or two false positives in each run, but it still learns much cleaner graphs compared to those learned by Reg.}
  \label{fig:causal_graph_manipulation_cmi}
  \vspace{-15pt}
\end{figure}

\newpage
\subsubsection{Predicting Future States Details}
\label{app:prediction_results}

In this section, we provide the results for 1 $\sim$ 5-step prediction results for each method, both evaluated on states in and out of training distribution (denoted as ID and OOD states) except for the physical environment for which it is not practical to create OOD states. To create OOD states, we change object positions in the chemical environment and marker positions in the manipulation environment to random values sampled from a normal distribution $\mathcal{N}(0, \sigma^2)$ where $\sigma$ controls to what extent the OOD states are different from ID states. Both ID and OOD prediction are measured on other unchanged state variables which do not depend on those changed variables, so ideally prediction performance on OOD states should be the same as the performance on ID states. For 1-step prediction results shown in Sec. \ref{subsubsec:prediction_results}, we use $\sigma=1000$ for the chemical environment and $\sigma=1$ for the manipulation environment.

\begin{figure}
  \centering
  \includegraphics[width=1.0\columnwidth]{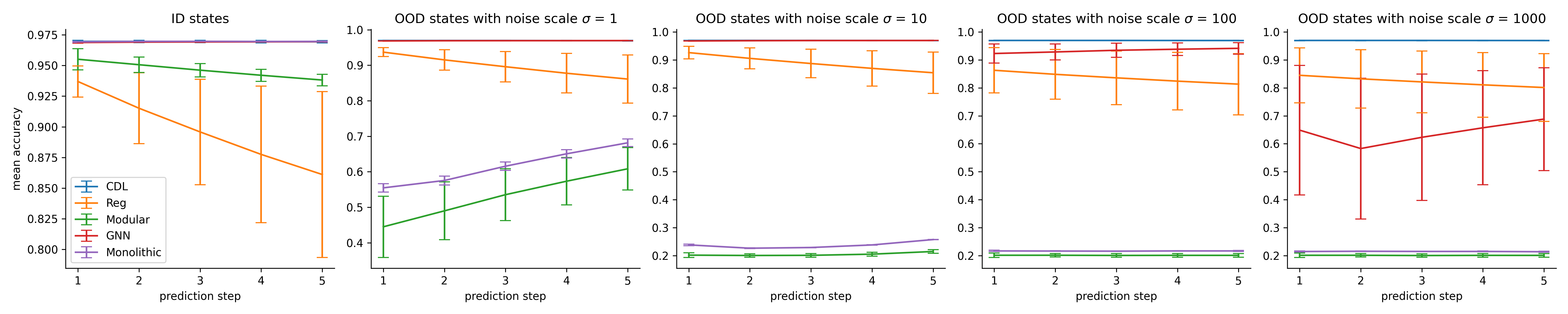}
  \vspace{-20pt}
  \caption{\small Multi-step prediction performance for the chemical environment with the \textbf{collider} graph. \textbf{(leftmost)} prediction on ID states. \textbf{(others)} prediction on OOD states with increasing noise scale $\sigma$.}
  \label{fig:multistep_prediction_collider}
  \vspace{-0pt}
\end{figure}

\begin{figure}
  \centering
  \includegraphics[width=1.0\columnwidth]{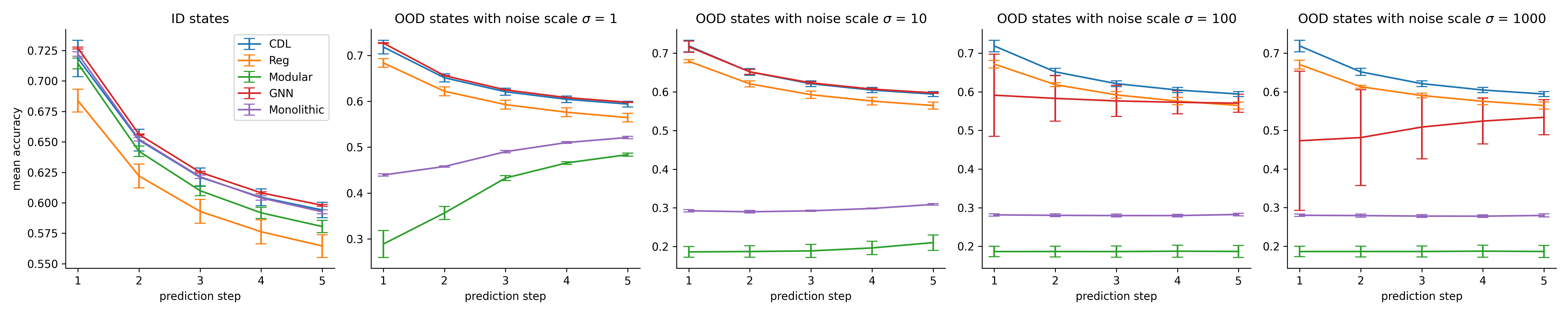}
  \vspace{-20pt}
  \caption{\small Multi-step prediction performance for the chemical environment with the \textbf{chain} graph. \textbf{(leftmost)} prediction on ID states. \textbf{(others)} prediction on OOD states with increasing noise scale $\sigma$.}
  \label{fig:multistep_prediction_chain}
  \vspace{-0pt}
\end{figure}

\begin{figure}
  \centering
  \includegraphics[width=1.0\columnwidth]{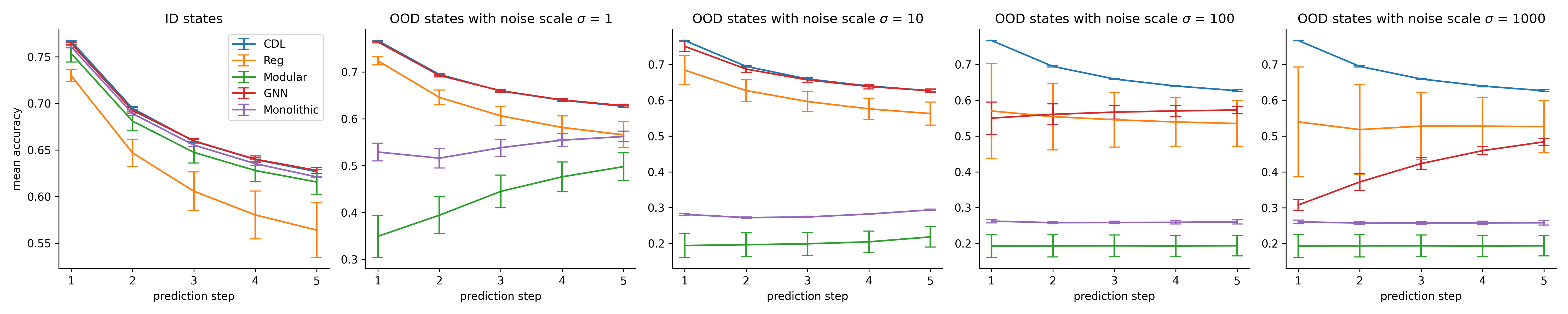}
  \vspace{-20pt}
  \caption{\small Multi-step prediction performance for the chemical environment with the \textbf{full} graph. \textbf{(leftmost)} prediction on ID states. \textbf{(others)} prediction on OOD states with increasing noise scale $\sigma$.}
  \label{fig:multistep_prediction_dense}
  \vspace{-0pt}
\end{figure}

\begin{figure}
  \centering
  \includegraphics[width=0.6\linewidth]{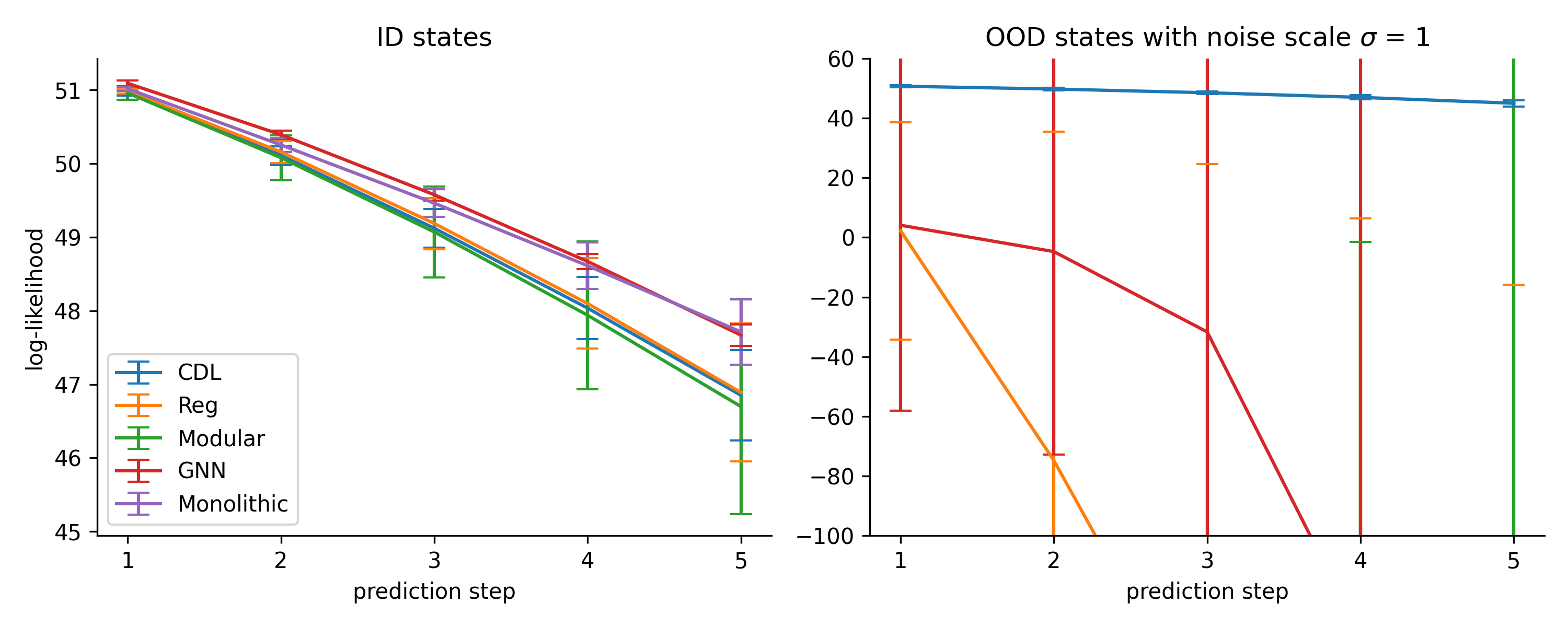}
  \vspace{-10pt}
  \caption{\small Multi-step prediction performance for the manipulation environment. \textbf{(leftmost)} prediction on ID states. \textbf{(others)} prediction on OOD states with noise scale $\sigma=1$.}
  \label{fig:multistep_prediction_manipulation}
  \vspace{-0pt}
\end{figure}

\begin{figure}
  \centering
  \includegraphics[width=0.4\linewidth]{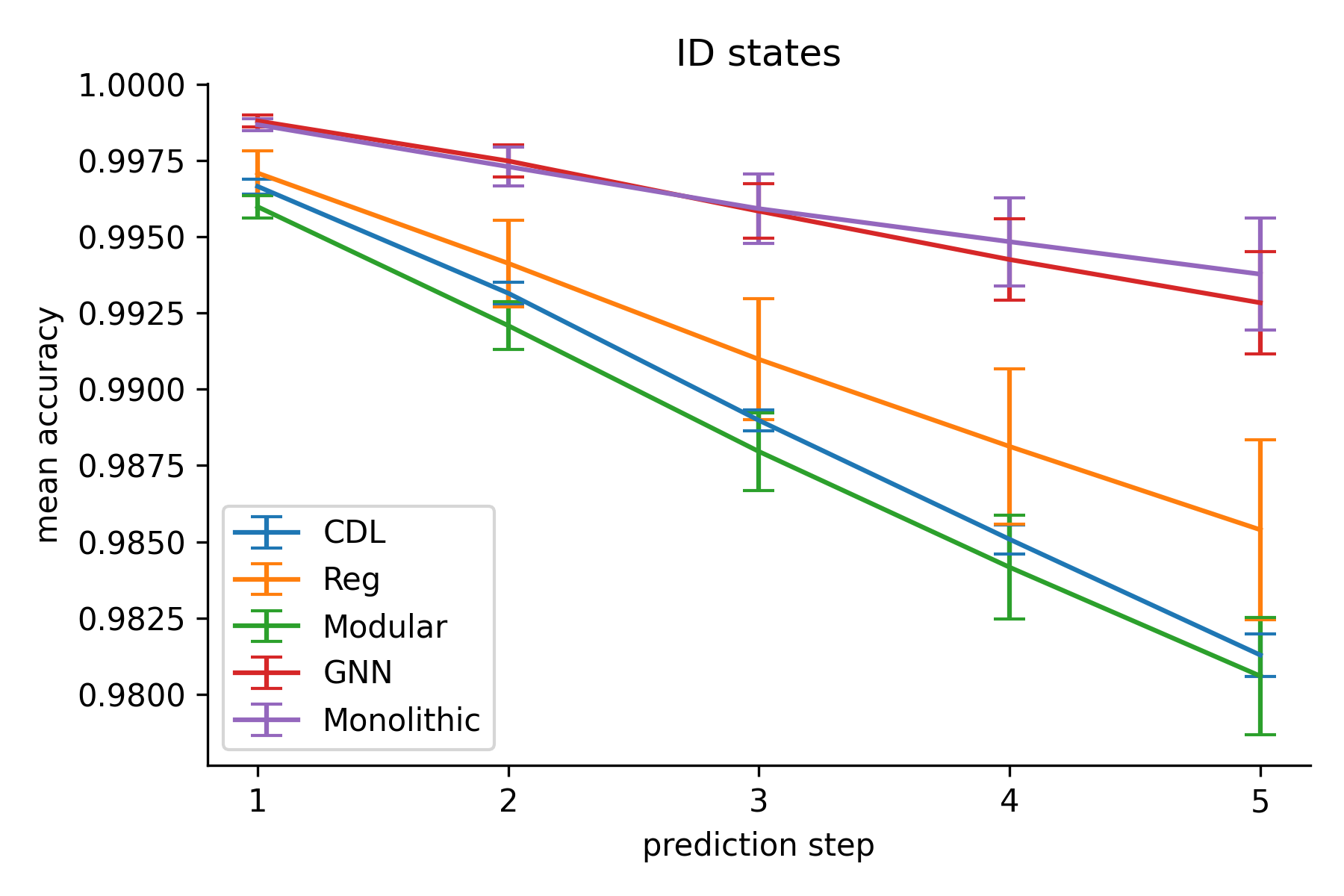}
  \vspace{-10pt}
  \caption[width=0.9\linewidth]{\small Multi-step prediction performance for the physical environment.}
  \label{fig:multistep_prediction_physical}
\end{figure}

The 1 $\sim$ 5-step prediction results are shown in Fig. \ref{fig:multistep_prediction_collider} $\sim$ \ref{fig:multistep_prediction_physical}. In all environments, \textsc{cdl} achieves similar performance to other methods on ID states. On OOD states, \textsc{cdl} retains similar performance while the performance of dense methods degrades severely. One exception is in the manipulation environment where one run of \textsc{cdl} wrongly uses the spurious correlation on the marker, as shown in Fig. \ref{fig:causal_graph_manipulation_cmi}, and thus its prediction performance also degrades a little. Meanwhile, although Reg also learns a causal graph and shares the generalization benefits, as it does not learn the causal graphs as accurately as \textsc{cdl}, its prediction performance is worse than \textsc{cdl} both on ID and OOD states. 

We also evaluate how much the prediction performance of dense model decreases under different $\sigma$ values, as shown in Fig. \ref{fig:multistep_prediction_collider} $\sim$ \ref{fig:multistep_prediction_dense}. Surprisingly, though GNN assumes a dense underlying graph between state variables, it still performs comparably with \textsc{cdl} when $\sigma=100$ in the chemical environment with the collider graph and when $\sigma=10$ in the chemical environment with the chain or dense graph. This suggests that the dense model has certain generalizability when being used in environments with simple underlying causal graphs (like the collider graph) and being trained with enough data. However, this is barely the case for more complex environments. For example, in the manipulation environment, GNN's prediction degrades severely even with small noise $\sigma=1$. Furthermore, we find that, counter-intuitively, for certain dense models, environments, and noise scales, the prediction performance increases with larger prediction steps, such as Modular and Monolithic in Fig. \ref{fig:multistep_prediction_collider} with $\sigma=1$. We hypothesize that this increasing performance results from the generalizability of dense models just described: though the state given to the models is OOD, as the model rolls out to predict longer into the future, it prefers to predict ID values for states and gradually recovers from the interruption brought by the OOD value. As mentioned, the generalizability of dense models is limited, so for larger values of $\sigma$, Modular and Monolithic cannot recover any more and thus the prediction performance no longer increases with the prediction step.

\subsection{Transition Collection Policy Learning Evaluation Details}
\label{app:data_collection_policy_results}

Apart from the results shown in Sec. \ref{subsec:data_collection_policy_results}, we provide other quantitative and qualitative results of the transition collection policy evaluation in this section.

\begin{table}
\centering
\caption{Performance on Causal Graph Learning for \textsc{cdl} and Reg on All Environments}
\vspace{-0pt}
\begin{small}
\begin{tabular}{cccc}
\toprule
\textbf{Metrics}    & \textbf{PredDiff}             & \textbf{Uniform}      & \textbf{Curiosity}     \\
\midrule
\textbf{Accuracy}   & \textbf{0.902} $\pm$ 0.003    & 0.891 $\pm$ 0.006     & 0.889 $\pm$ 0.002 \\
\textbf{Recall}     & 0.612 $\pm$ 0.009             & 0.593 $\pm$ 0.042     & 0.573 $\pm$ 0.009 \\
\textbf{Precision}  & 0.980 $\pm$ 0.006             & 0.943 $\pm$ 0.022     & 0.956 $\pm$ 0.022 \\
\textbf{F1 Score}   & 0.754 $\pm$ 0.008             & 0.726 $\pm$ 0.026     & 0.716 $\pm$ 0.005 \\
\textbf{ROC AUC}    & 0.805 $\pm$ 0.005             & 0.792 $\pm$ 0.019     & 0.784 $\pm$ 0.003 \\
\bottomrule
\end{tabular}
\end{small}
\vspace{-0pt}
\label{tab:transition_policy_results_app}
\end{table}

\begin{figure}
  \centering
  \includegraphics[width=1.0\columnwidth]{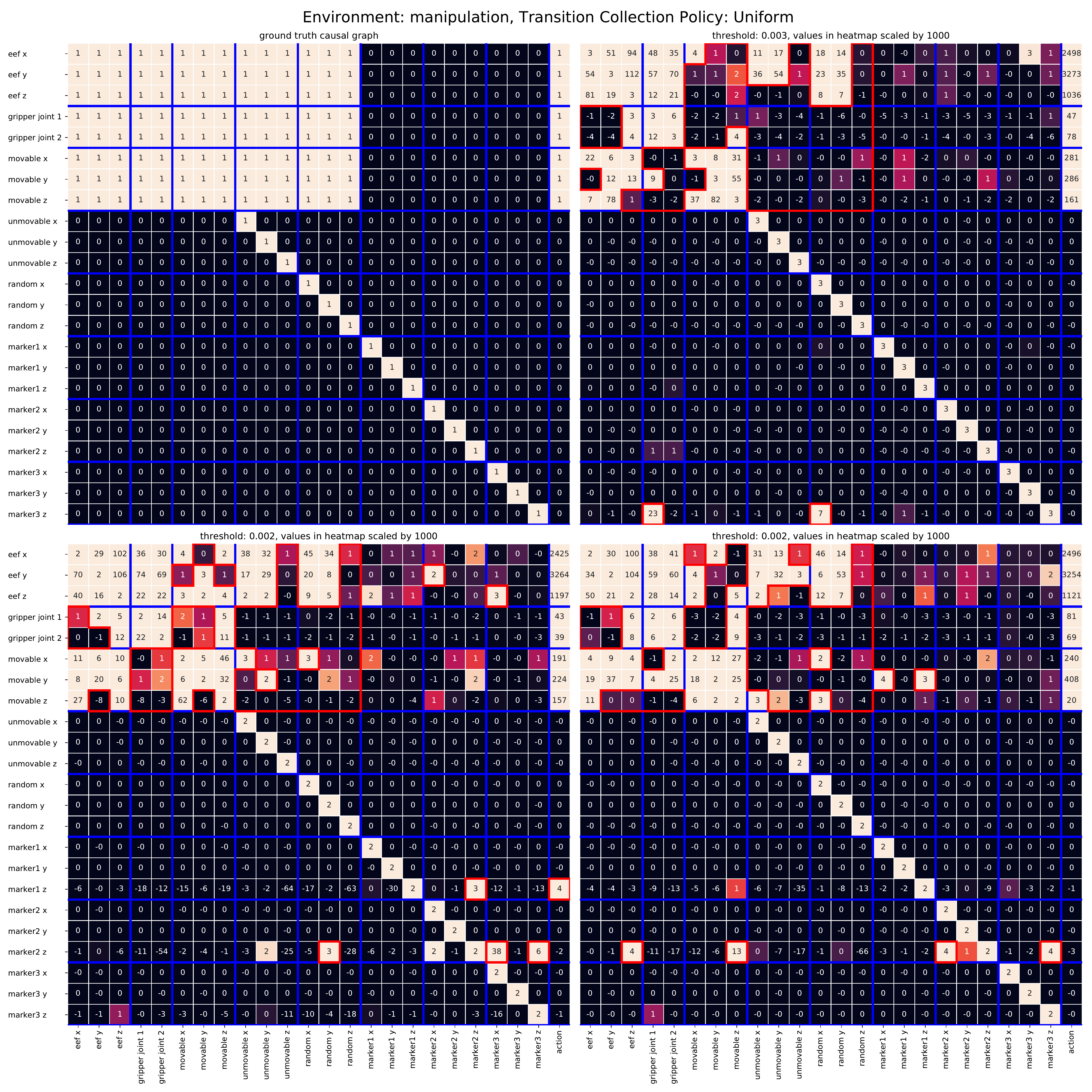}
  \vspace{-20pt}
  \caption{\small Causal graph for the manipulation environment learned with \textsc{cdl} and transitions collected by \textbf{Uniform}. \textbf{(top left)} the ground truth causal graph, \textbf{(others)} causal graphs learned with data from \textbf{Uniform} for 3 different runs. Compared to the graphs learned with data from \textbf{PredDiff} shown in Fig. \ref{fig:causal_graph_manipulation_cmi}, there are more spurious correlations (which is also reflected by the lower precision scores), especially in two bottom graphs, and those spurious correlations limit the generalizability of the learned causal dynamics models.}
  \label{fig:causal_graph_manipulation_uniform}
  \vspace{-15pt}
\end{figure}

\begin{figure}
  \centering
  \includegraphics[width=1.0\columnwidth]{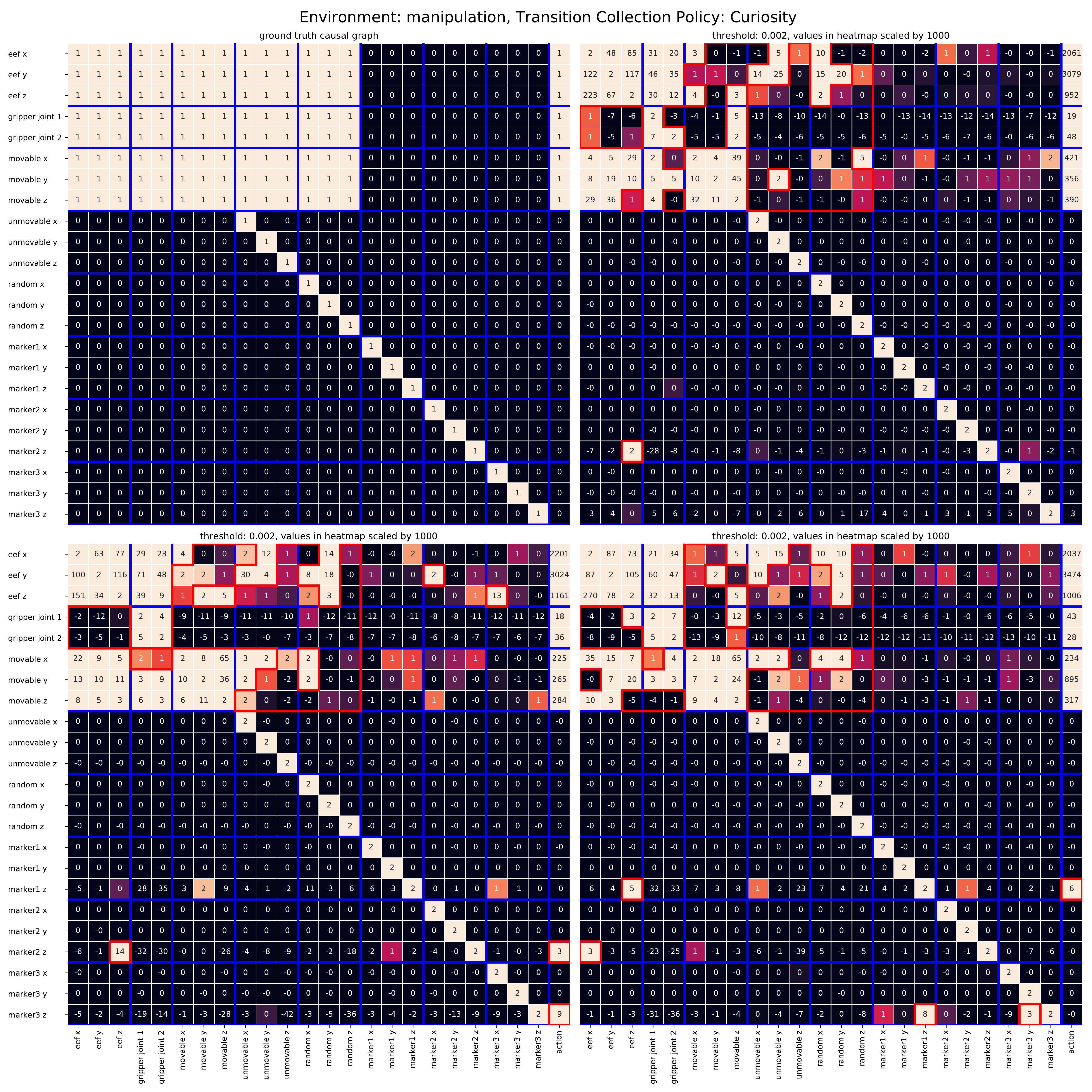}
  \vspace{-20pt}
  \caption{\small Causal graph for the manipulation environment learned with \textsc{cdl} and transitions collected by \textbf{Curiosity}. \textbf{(top left)} the ground truth causal graph, \textbf{(others)} causal graphs learned with data from \textbf{Curiosity} for 3 different runs. Compared to the graphs learned with data from \textbf{PredDiff} shown in Fig. \ref{fig:causal_graph_manipulation_cmi}, there are more missing dependencies (which is also reflected by the lower recall scores), especially eef's dependencies on the unmovable and randomly moving object. Those missing dependencies result from procrastination, a common problem for curiosity agents.}
  \label{fig:causal_graph_manipulation_curiosity}
  \vspace{-15pt}
\end{figure}

For the causal graph derivation, we use the same method described in Sec. \ref{app:causal_graph_results} for threshold selection and the same metrics to evaluate the causal graphs learned from data collected by different policies, which are shown in Table. \ref{tab:transition_policy_results_app}. The causal graphs learned by PredDiff is the same as Fig. \ref{fig:causal_graph_manipulation_cmi} and the ones learned by Uniform and Curiosity are shown in Fig, \ref{fig:causal_graph_manipulation_uniform} and Fig. \ref{fig:causal_graph_manipulation_curiosity} respectively. For the Curiosity policy, we also observe procrastination which is a common problem for curiosity-based exploration. Specifically, as it is harder to predict the next time step value of the movable object when it is being moved around than when it stays still, the agent keeps moving the movable object to maximize rewards, rather than actively exploring and exposing unlearned causal relationships, e.g., interacting with the unmovable and randomly moving object to expose eef's dependencies on them.

\subsection{Downstream Tasks Learning Evaluation Details}
\label{app:task_results}

In this section, we give more details on task setup, reward predictor implementations and downstream task learning results.

For the tasks described in Sec. \ref{subsec:task_results}, their reward functions are defined as follows:

\textbf{Matching} (C): match the object colors with goal colors individually,
\begin{equation}
    r_t = \sum_{i=1}^{10} \mathds{1}\left[ c_t^i = g^i \right], \nonumber
\end{equation}
where $\mathds{1}$ is the indicator function, $c_T^i$ is the current color of the $i$-th object, and $g^i$ is the goal color of the $i$-th object.

\textbf{Reach} (M): move the end-effector to the goal position,
\begin{equation}
    r_t = \tanh(10 \cdot \lVert eef_t - g\rVert_1), \nonumber
\end{equation}
where $\lVert \cdot \rVert_1$ is L1 norm, $eef_t \in \mathbb{R}^3$ is the current end-effector position, and $g \in \mathbb{R}^3$ is the goal position in this episode.

\textbf{Lift} (M): lift the movable object to the goal position,
\begin{equation}
r_t = 
\begin{cases}
0.3 + 0.5 \cdot (1  - \frac{\lVert mov_t - g\rVert_1}{1.2})                                                         & \text{if $mov$ is grasped}\\
0.1 \cdot (1 - \frac{\lVert eef_t - mov_t\rVert_1}{1.2}) \cdot \mathds{1}\left[\text{gripper is open} \right]       & \text{otherwise}
\end{cases}.
\nonumber
\end{equation}

\textbf{Stack} (M): stack the movable object on the top of the unmovable object.
\begin{equation}
r_t = 
\begin{cases}
0.35 + 1.0 \cdot (1  - \frac{\lVert mov_t - g\rVert_1}{1.2})                                                             & \text{if $mov$ is grasped}\\
2.0                                                                                                                 & \text{if $mov$ is not grasped and is on top of $unm$} \\
0.1 \cdot (1 - \frac{\lVert eef_t - mov_t\rVert_1}{1.2}) \cdot \mathds{1}\left[\text{gripper is open} \right]       & \text{otherwise}
\end{cases}.
\nonumber
\end{equation}

\begin{table}
\centering
\caption{Parameters of the Reward Predictor and \textsc{cem} (Shared Across Tasks if not Specified)}
\vspace{-0pt}
\begin{small}
\begin{tabular}{ccccccc}
\toprule
\textbf{Method} & \textbf{Name}                 & \multicolumn{5}{c}{\textbf{Tasks}} \\
                &                               & \thead{Chemical\\(collider)}  
                                                & \thead{Chemical\\(chain, full)}  
                                                & Reach     & Manipulation  & Physical    \\
\midrule
\multirow{6}{*}{\thead{Reward\\Predictor}}
& architecture                      & \multicolumn{2}{c}{[64, 64]}      & [128, 64]     & \multicolumn{2}{c}{[64, 64, 64]}          \\
& activation functions              & \multicolumn{2}{c}{[ReLU, ReLU]}  & [ReLU, Tanh]  & \multicolumn{2}{c}{[ReLU, ReLU, ReLU]}    \\
& training step                     & 300K      & 1M        & 2M        & 2M        & 2M    \\
& optimizer                         & \multicolumn{5}{c}{Adam}                              \\
& learning rate                     & \multicolumn{5}{c}{3e-4}                              \\
& batch size                        & \multicolumn{5}{c}{32}                            \\
\midrule
\multirow{6}{*}{\textsc{cem}}
& planning length, $L$              & \multicolumn{2}{c}{3}             & \multicolumn{3}{c}{1}     \\
& number of candidates, $J$         & \multicolumn{2}{c}{64}            & \multicolumn{3}{c}{128}   \\
& number of top candidates, $K$     & \multicolumn{2}{c}{32}            & \multicolumn{3}{c}{32}    \\
& number of iterations, $N$         & \multicolumn{2}{c}{5}             & \multicolumn{3}{c}{10}    \\
& exploration noise                 & \multicolumn{2}{c}{N/A}           & \multicolumn{3}{c}{0.03}  \\
& exploration probability           & \multicolumn{2}{c}{0.2}           & \multicolumn{3}{c}{N/A}   \\
\bottomrule
\end{tabular}
\end{small}
\vspace{-0pt}
\label{tab:task_policy_parameters}
\end{table}

During training, the dynamics model is frozen and only the reward predictor is learned as described in Sec. \ref{subsec:task_learning}. The architecture and hyperparameters of the reward predictor are listed in Table. \ref{tab:task_policy_parameters}, along with the parameters of Cross-Entropy Method (\textsc{cem}) that is used for planning.

\begin{figure}
  \centering
  \includegraphics[width=1.0\columnwidth]{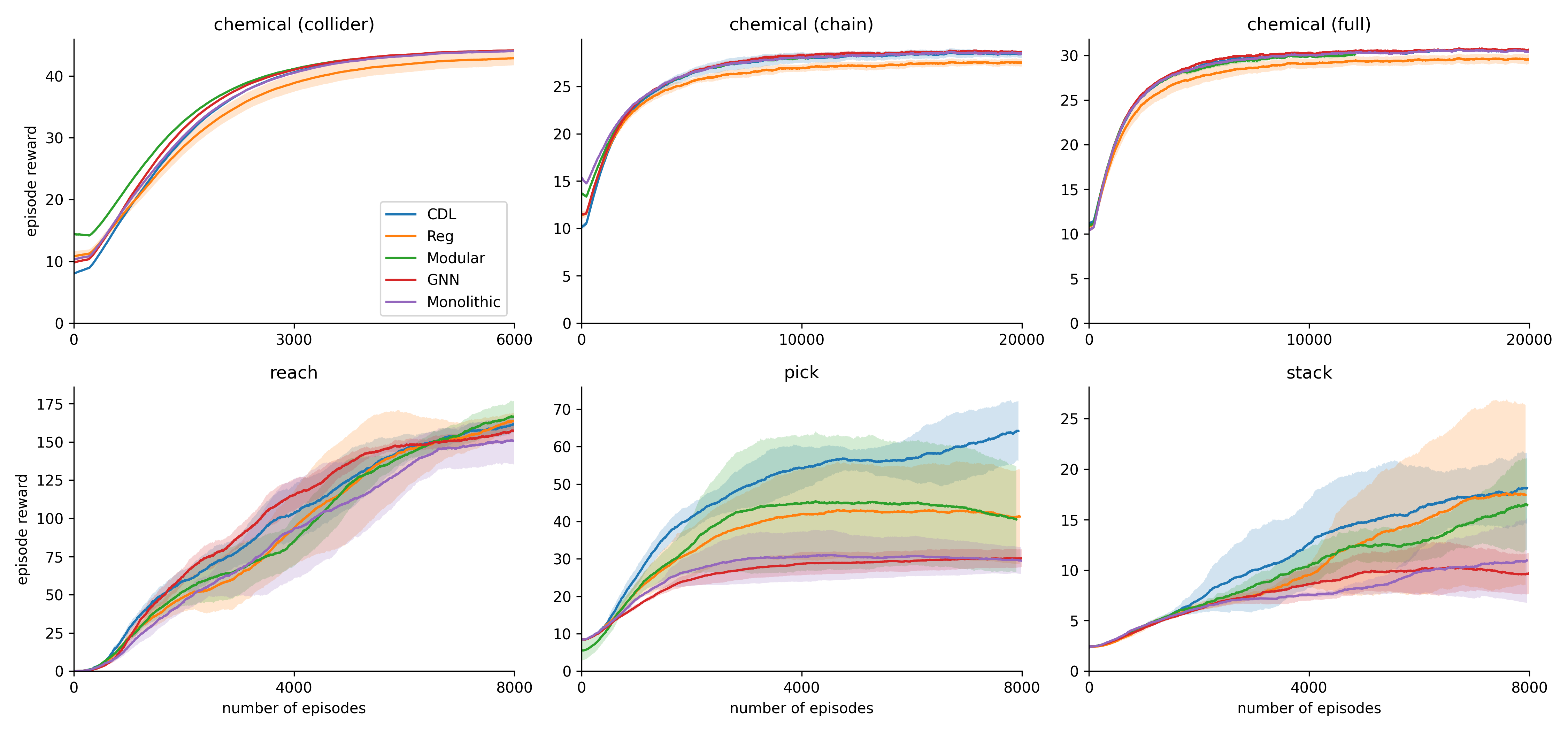}
  \vspace{-20pt}
  \caption{\small Traninig curve for all downstream tasks.}
  \label{fig:learning_curve_all}
  \vspace{-0pt}
\end{figure}

The training curves for all tasks are shown in Fig. \ref{fig:learning_curve_all}. Compared to lift and stack, the sample efficiency advantages of the derived are less obvious for match and reach tasks, because those tasks are relatively simple and have dense reward signals that are easy to learn.

\end{document}